\documentclass[11pt]{article}

\usepackage[margin=1in]{geometry}
\usepackage{graphicx}
\usepackage{amsmath,amssymb,amsfonts}
\usepackage{mathtools}
\usepackage{url}
\usepackage{booktabs}
\usepackage{enumitem}
\usepackage{float}
\usepackage{natbib}
\usepackage{algorithm}
\usepackage{algorithmicx}
\usepackage{algpseudocode}
\usepackage{tikz}
\usepackage{hyperref}
\usepackage{bm}

\usetikzlibrary{positioning,arrows.meta,shapes,calc}

\hypersetup{
  colorlinks=true,
  linkcolor=blue,
  citecolor=blue,
  urlcolor=blue
}

\newtheorem{theorem}{Theorem}
\newtheorem{proposition}{Proposition}

\newtheorem{assumption}{Assumption}

\newcommand{\T}{^\top}

\newcommand{\Halmos}{\hfill$\square$}
\providecommand{\BIBand}{and}

\title{Personalized Federated Hierarchical Gaussian Processes for Privacy-Preserving Modeling of Heterogeneous Distributed Systems}
\author{
  Xianjian Xie\thanks{Arizona State University. Email: \texttt{xxie43@asu.edu}.} \and
  Hao Yan\thanks{Arizona State University. Email: \texttt{haoyan@asu.edu}.}
}
\date{}

\begin{document}
\maketitle

\begin{abstract}
We present Personalized Federated Hierarchical Gaussian Processes (pFedHGP) for probabilistic regression and classification when data are distributed across heterogeneous clients. Each client's latent function decomposes into (i) a shared global component, (ii) a client-specific deviation that shares the global kernel structure, and (iii) a flexible local residual. Sparse inducing-variable approximations and federated variational inference keep raw data local while the server synchronizes only low-dimensional statistics for the shared component. Full predictive distributions support uncertainty-aware decisions. In application studies, pFedHGP attains perfect fault classification in press tonnage monitoring using 13.77\% of labeled cycles and recovers geographic zones in federated air-quality modeling without centralizing station-level time series. An Instantaneous Linear Mixing Model viewpoint links the hierarchy to multi-output Gaussian processes for correlated sensors.
\end{abstract}


\textbf{Funding:} This work is funded by DOE NEUP DE-NE0009383.

\section{Introduction}

Privacy-preserving learning from spatially and institutionally distributed measurements has become central to modern industrial and urban analytics \citep{kontar2021internet}. With the rapid proliferation of Internet of Things (IoT) devices---distributed environmental sensors, manufacturing equipment monitors, and smart city infrastructure---many deployments naturally adopt federated architectures where data remain on edge devices. In this setting, models must capture population-level regularities while respecting data locality, bandwidth limits, and heterogeneous operating conditions. Representative application domains include smart healthcare \citep{xu2021federated}, predictive maintenance in industrial manufacturing \citep{pruckovskaja2023federated}, and connected vehicle networks for smart cities \citep{chellapandi2023federated}.

A fundamental challenge in such distributed systems is handling heterogeneity: physical systems often exhibit site-specific behaviors due to differing operational conditions, environmental factors, equipment variations, or localized disturbances. When learning shared models from non-i.i.d.\ federated clients, this heterogeneity can significantly degrade prediction quality and yield poorly calibrated uncertainty \citep{ye2023heterogeneous}. The core tension is to retain a population-level representation of common dynamics while allowing client-specific systematic shifts and flexible local residuals.

To address heterogeneity, a stream of research in personalized learning, including personalized federated learning \citep{deng2020adaptive}, personalized PCA \citep{shi2024personalized}, and personalized Tucker decomposition \citep{hu2025personalized}, has focused on decomposing client data into shared global components and local personalized components. These approaches typically extract a joint low-rank subspace to represent the global structure while allowing client-specific variations in complementary subspaces. Essentially, they perform latent subspace decomposition by identifying shared bases along with client-specific coefficients, thereby achieving a balance between collective representation and individualized adaptation.

Inspired by this decomposition paradigm, our objective is to develop a federated probabilistic framework that separates shared structure from client-specific variation in settings with correlated multi-sensor outputs. Such scenarios arise in manufacturing systems with multiple strain channels, urban environmental monitoring networks, and other distributed sensing stacks. We leverage Gaussian processes (GPs), which naturally encode dependence through kernels, provide principled uncertainty quantification (UQ), and remain flexible in low-data regimes \citep{seeger2004gaussian}. Rigorous uncertainty estimates are particularly important for monitoring and planning under safety constraints. GPs have strong records in surrogate modeling, spatiotemporal forecasting, robotics, and sensor networks \citep{deisenroth2013gaussian, wilson2013gaussian}.

Despite existing efforts, three major challenges remain in developing GP-based models for federated heterogeneous environments:

\begin{itemize}
\item \textbf{Lack of explicit hierarchical calibration structure:} Many federated GP methods rely on a single shared kernel with local fine-tuning, without a transparent decomposition into shared global structure, structured client deviations, and flexible local residuals.
\item \textbf{Inadequate support for correlated multi-sensor clients:} Existing federated GP methods often treat outputs as independent or rely on global pooling, limiting explicit modeling of cross-sensor dependence within each client.
\item \textbf{Insufficient uncertainty-aware adaptation:} Many federated methods do not deliver calibrated uncertainty that balances global consistency with client-specific adaptation, which matters for risk-aware decisions under heterogeneity.
\end{itemize}

To address these gaps, we propose personalized Federated Hierarchical Gaussian Process (pFedHGP), trained with privacy-preserving federated variational inference. pFedHGP embeds a three-level hierarchical decomposition: (1) a global GP capturing population-level structure, (2) client-specific deviation GPs that share the global kernel geometry, and (3) client-specific local residual GPs with independent kernels and inducing geometries. An Instantaneous Linear Mixing Model (ILMM) viewpoint \citep{bruinsma2020scalable} clarifies how correlated multi-sensor outputs arise from latent mixing; see Section~\ref{sec:ilmm_formulation}.

We summarize our contributions below:
\begin{itemize}
\item \textbf{Hierarchical decomposition for heterogeneous federated clients:} We decompose each client's latent function into shared global structure, structured deviations on a shared basis, and flexible local residuals, enabling interpretable monitoring under non-i.i.d.\ clients.
\item \textbf{Privacy-preserving federated variational inference:} We develop a two-stage variational procedure in which the server synchronizes only compact global statistics while clients update local variational factors on private data.
\item \textbf{Uncertainty-aware multi-sensor extension:} Full Gaussian predictive laws support marginal and joint uncertainty across outputs; the ILMM connection highlights how vector-valued sensing maps to the hierarchy.
\end{itemize}

The federated formulation is needed for three reasons. First, raw sensor trajectories are naturally owned by local clients and may contain proprietary or governance-restricted information. Second, high-resolution time series create communication and storage costs when naively centralized. Third, the clients are statistically heterogeneous, so a purely pooled model can obscure local systematic shifts. pFedHGP addresses these issues by keeping raw observations local, transmitting only model-level updates for the shared representation, and preserving client-specific behavior through the deviation and local components. We use the term privacy-preserving in this data-locality sense; the present work does not provide formal differential privacy or secure aggregation guarantees.

The remainder of this paper is organized as follows. Section~\ref{sec:background} reviews related works. Section~\ref{sec:methodology} introduces pFedHGP. Section~\ref{sec:experiments} presents a multi-output synthetic study with ablations, robustness checks under assumption mismatch, and uncertainty metrics. Section~\ref{sec:case study} presents realistic case studies. Section~\ref{sec:limitations} discusses limitations. Section~\ref{sec:conclusion} concludes.
\section{Background and Literature Review}
\label{sec:background}

The evolution of Gaussian Processes (GPs) from centralized models to distributed and federated architectures has been driven by the dual imperatives of computational scalability and data privacy. We categorize the existing literature into five major domains: GP-based calibration and model-discrepancy methods (Section~\ref{sec:gp_calibration}), which assume centralized data access; distributed GP methods (Section~\ref{sec:distributed_gp}), which prioritize scalability over privacy; federated GP approaches (Section~\ref{sec:federated_gp}), which preserve privacy but struggle with client heterogeneity; personalized federated learning (Section~\ref{sec:personalized_fl}), which combines global and local adaptations; and advanced federated architectures (Section~\ref{sec:advanced_federated}), which handle multi-output and multi-fidelity data at the cost of increased complexity.

\subsection{Gaussian Processes for Calibration and Model Discrepancy}
\label{sec:gp_calibration}
Gaussian Processes provide a probabilistic framework for model calibration, uncertainty quantification, and correction of discrepancies between simulation models and physical observations. \citet{chakraborty2020role} highlight the role of GP surrogates in bridging the gap between high-fidelity simulations and noisy sensor data. In structural dynamics, \citet{kessels2022model} use GPs as inverse mapping models to update physical parameters and quantify calibration uncertainty. These centralized approaches leverage the flexibility of GPs to model the systematic bias, or model-discrepancy, function \citep{kennedy2001bayesian} that captures the difference between an idealized physical model and observed data. However, traditional calibration methods typically assume centralized data access, which becomes problematic in distributed industrial settings where data privacy is paramount.

\subsection{Distributed Gaussian Processes}
\label{sec:distributed_gp}
Early distributed GP methods were primarily designed to address the cubic computational complexity of standard GPs \citep{kontoudis2022fully, liu2018generalized}. These approaches typically partition data across multiple nodes to parallelize computation. \citet{tresp2000bayesian} introduced the Bayesian Committee Machine (BCM), which aggregates predictions from independent GP experts by weighing them with their inverse covariance. \citet{deisenroth2015distributed} extended this to a distributed variational framework (DgGP), allowing for the robust aggregation of expert predictions. Similarly, \citet{ng2014hierarchical} and \citet{nguyen2014fast} proposed mixture-of-experts (MoE) models that assign data to local experts based on hierarchical or information-theoretic criteria. 

In signal processing, \citet{xu2019distributed} and \citet{yin2017distributed} developed distributed recursive algorithms for wireless traffic and sensor networks, focusing on efficient hyperparameter synchronization. \citet{peng2017asynchronous} and \citet{gal2014distributed} further improved scalability by leveraging asynchronous updates and stochastic variational inference, enabling GPs to handle massive datasets.

The primary limitation of distributed GPs is their assumption that data is either i.i.d. or artificially partitioned for computational convenience. They often rely on shared global kernels or inducing points, lacking mechanisms to model the intrinsic heterogeneity (non-i.i.d. distributions) found in natural federated settings. Furthermore, these methods prioritize scalability over privacy, often requiring the exchange of gradients or statistics that may leak sensitive information \citep{yin2020fedloc}.

\subsection{Federated Gaussian Processes}
\label{sec:federated_gp}
With the rise of privacy concerns, the focus shifted to Federated Learning (FL), where data remains local. Federated GPs aim to learn a global GP model without sharing raw data. \citet{yu2022federated} proposed Federated Bayesian Neural Regression (FedBNR), utilizing random features to approximate a scalable global kernel. \citet{yurochkin2019bayesian} introduced Bayesian Nonparametric Federated Learning (BNFed) to match neural network weights across clients probabilistically.

Recent works have explored decentralized approaches. \citet{kontoudis2022fully} developed fully decentralized GPs for multi-agent systems, while \citet{llorente2025robust} introduced robust, online decentralized GPs adaptable to dynamic environments. \citet{thorgeirsson2020probabilistic} provided a probabilistic extension to the standard FedAvg algorithm, aggregating moments of client distributions to quantify global uncertainty.

While these methods improve privacy, they generally enforce a single global model across all clients. This "one-size-fits-all" approach struggles when clients exhibit significant statistical heterogeneity. A single global kernel cannot capture the diverse, client-specific patterns inherent in many real-world applications, leading to suboptimal performance for individual participants.

\subsection{Personalized Federated Learning}
\label{sec:personalized_fl}
To address heterogeneity, Personalized Federated Learning (PFL) integrates global knowledge with client-specific adaptations. While Gaussian processes offer one avenue for personalization, broader PFL frameworks have explored various decomposition strategies. \citet{shi2024personalized} introduced Personalized PCA (PerPCA), which explicitly decouples client data into shared global features and unique local features, enabling robust dimensionality reduction in heterogeneous settings. Similarly, \citet{hu2025personalized} proposed Personalized Tucker Decomposition (perTucker) for tensor data, modeling commonalities via a shared core tensor while capturing peculiarities through client-specific factor matrices.

In the context of GPs, the seminal work on pFedGP by \citet{achituve2021personalized} combines Deep Kernel Learning (DKL) with local GPs. In this framework, a neural network feature extractor (the deep kernel) is learned globally, while each client maintains a local GP on the extracted features. This allows for shared representation learning while preserving local flexibility. Subsequent research has refined this paradigm; for instance, \citet{yang2024efficient} incorporated reinforcement learning to optimize personalization strategies in heterogeneous environments.

Non-GP personalized methods like PerPCA and perTucker are powerful for feature extraction but lack the built-in uncertainty quantification needed for safety-critical monitoring and decision-making tasks. On the other hand, personalized federated GPs like pFedGP often suffer from high computational burdens on client devices ($O(N^3)$ complexity) and communication overheads associated with deep kernel parameters.

\subsection{Advanced Federated Architectures}
\label{sec:advanced_federated}
Recent advancements have extended federated GPs to handle more complex data structures, such as multi-output and multi-fidelity data. \citet{chung2024federated} introduced Federated Multi-output Gaussian Processes (FedMGP), which learn shared latent functions to model correlations across multiple outputs (e.g., spatial or temporal tasks). \citet{gao2024federated} further enhanced this by developing automated mechanisms for selecting latent variables in federated settings.

In reliability engineering, \citet{jeong2025fed} proposed Fed-Joint, a framework coupling nonlinear degradation signals with failure events using federated multi-output GPs. Similarly, \citet{yue2024federated} explored federated data analytics for linear models, providing a foundation for scalable multi-fidelity modeling.

These advanced architectures introduce significant structural complexity. They often require sophisticated variational inference schemes \citep{bruinsma2020scalable} to remain tractable, and the need to capture correlations across multiple outputs or fidelities increases the communication overhead and the difficulty of optimizing the global variational objective.

\section{Methodology}
\label{sec:methodology}

We present the Personalized Federated Hierarchical Gaussian Process (pFedHGP) framework for privacy-preserving federated regression and calibration under client heterogeneity. Section~\ref{sec:preliminaries} reviews sparse Gaussian processes and the Variational Free Energy framework that enables scalable inference. Section~\ref{sec:motivation_and_formulation} introduces the hierarchical generative model decomposing observations into global, client-specific deviation, and local residual components. Section~\ref{sec:inducing_variables} establishes the low-rank inducing variable representation that provides computational efficiency. Section~\ref{sec:variational_sparse_assumption} presents the variational sparse GP framework with mean-field assumptions and additive covariance decomposition. Section~\ref{sec:variational_inference} derives the federated variational inference algorithm with explicit ELBO decomposition. Section~\ref{sec:algorithm} details the pFedHGP training algorithm alternating between local optimization and global aggregation. Section~\ref{sec:implementation} covers implementation details including Cholesky factorization, whitened parameterization, and hyperparameter optimization. Section~\ref{sec:usage} describes practical usage for classification and clustering tasks. Finally, Section~\ref{sec:ilmm_formulation} connects pFedHGP to the Instantaneous Linear Mixing Model framework.

\subsection{Preliminaries: Sparse Gaussian Processes}
\label{sec:preliminaries}

Standard Gaussian processes (GPs) suffer from cubic computational complexity $O(N^3)$, making them unsuitable for large-scale or federated learning. Sparse Gaussian processes (SGPs) address this by introducing a set of $M \ll N$ inducing variables $\mathbf{u}$ at locations $\mathbf{Z} = [\mathbf{z}_1, \dots, \mathbf{z}_M]$. For a single GP with prior $f \sim \mathcal{GP}(0,k)$, the joint prior over function values $\mathbf{f} = f(\mathbf{X})$ and inducing values $\mathbf{u} = f(\mathbf{Z})$ is Gaussian, and the conditional distribution
\begin{equation}
p(\mathbf{f} \mid \mathbf{u}) 
= \mathcal{N}\Big(\mathbf{K}_{nm}\mathbf{K}_{mm}^{-1}\mathbf{u},\;
 \mathbf{K}_{nn} - \mathbf{K}_{nm}\mathbf{K}_{mm}^{-1}\mathbf{K}_{mn}\Big)
\end{equation}
is \emph{exact}. Here $\mathbf{K}_{nn}=k(\mathbf{X},\mathbf{X})$, $\mathbf{K}_{mm}=k(\mathbf{Z},\mathbf{Z})$, and $\mathbf{K}_{nm}=k(\mathbf{X},\mathbf{Z})$.

Titsias's variational formulation views the inducing variables as variational parameters. Introducing a variational distribution of the form
\[
\begin{aligned}
q(\mathbf{f},\mathbf{u}) = p(\mathbf{f}\mid\mathbf{u})\,q(\mathbf{u}),\quad
q(\mathbf{u}) = \mathcal{N}(\mathbf{m},\mathbf{S}),
\end{aligned}
\]
yields the evidence lower bound (ELBO) under the Variational Free Energy (VFE) framework.

\begin{proposition}[Sparse GP ELBO]
\label{prop:vfe_elbo}
Under the VFE framework with inducing variables $\mathbf{u}$ at locations $\mathbf{Z}$, the ELBO takes the form:
\begin{equation}
\label{eq:sgp_vfe}
\mathcal{L}_{\mathrm{VFE}}
= \mathbb{E}_{q(\mathbf{u})}\big[\log p(\mathbf{y}\mid \mathbf{u})\big]
- \mathrm{KL}\big(q(\mathbf{u})\Vert p(\mathbf{u})\big)
- \frac{1}{2\sigma^2}\operatorname{Tr}\big(\mathbf{K}_{nn}-\mathbf{Q}_{nn}\big),
\end{equation}
where $p(\mathbf{y}\mid\mathbf{u}) = \mathcal{N}(\mathbf{K}_{nm}\mathbf{K}_{mm}^{-1}\mathbf{u}, \sigma^2\mathbf{I})$,
$\mathbf{Q}_{nn} = \mathbf{K}_{nm}\mathbf{K}_{mm}^{-1}\mathbf{K}_{mn}$, and $\sigma^2$ is the observation noise variance. The first two terms correspond to a low-rank approximating model with covariance $\mathbf{Q}_{nn}+\sigma^2\mathbf{I}$, and the trace term penalizes the discrepancy between the full prior covariance $\mathbf{K}_{nn}$ and its low-rank approximation.
\end{proposition}

The VFE method is originally proposed in \citep{titsias2009variational}, and 
a proof of Proposition~\ref{prop:vfe_elbo} is provided in Appendix~\ref{app:vfe_proof}.

The key viewpoint we adopt is that SGPs are not defined by modifying the GP prior, but by introducing inducing variables and optimizing a variational bound of the form~\eqref{eq:sgp_vfe}. Our pFedHGP method extends this variational sparse GP view to a federated setting by decomposing the latent function space into shared and local components, each equipped with its own inducing variables.


\subsection{Motivation and Formulation}
\label{sec:motivation_and_formulation}

For clarity, Table~\ref{tab:notation} summarizes the main notation used throughout the methodology.

\begin{table}[ht]
\centering
\caption{Summary of main notation used in pFedHGP.}
\label{tab:notation}
\small
\renewcommand{\arraystretch}{0.88}
\begin{tabular}{lll}
\toprule
Symbol & Domain & Description \\
\midrule
\(T\) & scalar & Number of clients \\
\(n_i\) & scalar & Number of observations at client \(i\) \\
\(\mathbf{X}_i\) & \(\mathbb{R}^{n_i\times d}\) & Inputs at client \(i\) \\
\(\mathbf{y}_i\) & \(\mathbb{R}^{n_i}\) & Observations at client \(i\) \\
\(f_g\) & function & Shared global latent function \\
\(f_{\delta,i}\) & function & Client-specific deviation function \\
\(f_i\) & function & Client-specific local residual function \\
\(k_g\) & kernel & Kernel of \(f_g\) \\
\(k_{\delta}\) & kernel & Kernel of \(f_{\delta,i}\) \\
\(k_i\) & kernel & Kernel of \(f_i\) \\
\(\phi\) & scalar & Deviation-kernel scaling parameter \\
\(\mathbf{Z}_g\) & \(\mathbb{R}^{M\times d}\) & Global inducing inputs \\
\(\mathbf{Z}_i\) & \(\mathbb{R}^{M_i\times d}\) & Local inducing inputs at client \(i\) \\
\(\mathbf{u}_g\) & \(\mathbb{R}^{M}\) & Global inducing variables \\
\(\boldsymbol{\delta}_i\) & \(\mathbb{R}^{M}\) & Deviation variables on \(\mathbf{Z}_g\) \\
\(\mathbf{u}_i\) & \(\mathbb{R}^{M_i}\) & Local inducing variables on \(\mathbf{Z}_i\) \\
\(\mathbf{A}_{g,i}\) & \(\mathbb{R}^{n_i\times M}\) & Global interpolation: \(\mathbf{K}_g(\mathbf{X}_i,\mathbf{Z}_g)\mathbf{K}_{gg}^{-1}\) \\
\(\mathbf{A}_{\delta,i}\) & \(\mathbb{R}^{n_i\times M}\) & Deviation interpolation (\(k_\delta=\phi k_g\)): \(\mathbf{A}_{g,i}\) \\
\(\mathbf{A}_{i}\) & \(\mathbb{R}^{n_i\times M_i}\) & Local interpolation: \(\mathbf{K}_i(\mathbf{X}_i,\mathbf{Z}_i)\mathbf{K}_{ii}^{-1}\) \\
\(B_g\) & \(\mathbb{R}^{Q\times R_g}\) & ILMM global loading (multi-output; Section~\ref{sec:ilmm_formulation}) \\
\(B_{\delta,i}\) & \(\mathbb{R}^{Q\times R_\delta}\) & ILMM deviation loading at client \(i\) \\
\(B_{l,i}\) & \(\mathbb{R}^{Q\times R_l}\) & ILMM local loading at client \(i\) \\
\(\mathbf{K}_{gg}\) & \(\mathbb{R}^{M\times M}\) & Gram matrix \(k_g(\mathbf{Z}_g,\mathbf{Z}_g)\) \\
\(\mathbf{K}_{ii}\) & \(\mathbb{R}^{M_i\times M_i}\) & Gram matrix \(k_i(\mathbf{Z}_i,\mathbf{Z}_i)\) \\
\bottomrule
\end{tabular}
\renewcommand{\arraystretch}{1.0}
\end{table}

We consider a federated learning scenario involving \(T\) clients, each of which holds private data inaccessible to a central server. Let client \(i\) possess a local dataset \(\{(\mathbf{x}_{i,j}, y_{i,j})\}_{j=1}^{n_i}\), where \(\mathbf{x}_{i,j} \in \mathbb{R}^d\) is the input and \(y_{i,j} \in \mathbb{R}\) is the corresponding scalar response. We assume that each observation is composed of three latent components---a shared global function, a client-specific deviation, and a local residual---plus additive Gaussian noise.

\begin{assumption}[pFedHGP Generative Model]
\label{assump:generative_model}
The observation $y_{i,j}$ for client $i$ is decomposed as
\begin{equation}
\label{eq:obs_decomposition}
y_{i,j} = f_g(\mathbf{x}_{i,j}) + f_{\delta,i}(\mathbf{x}_{i,j}) + f_i(\mathbf{x}_{i,j}) + \varepsilon_{i,j},
\end{equation}
where $f_g$ is the shared global function, $f_{\delta,i}$ is the client-specific deviation from the global trend, and $f_i$ is the local site-specific variation. The noise variables are i.i.d.\ $\varepsilon_{i,j}\sim\mathcal{N}(0,\sigma^2)$.
We endow these components with independent GP priors:
\begin{equation}
f_g \sim \mathcal{GP}\big(0, k_g\big),\quad
f_{\delta,i} \sim \mathcal{GP}\big(0, k_\delta\big),\quad
f_i \sim \mathcal{GP}\big(0, k_i\big),\quad i=1,\dots,T,
\end{equation}
with $f_g$, $\{f_{\delta,i}\}$, and $\{f_i\}$ mutually independent. In our implementation we take $k_\delta = \phi\,k_g$ so that the deviation processes share the same structure as the global process but are scaled by a factor $\phi>0$.
\end{assumption}

The three latent components have distinct statistical roles. The global process $f_g$ represents the population-level structure shared across clients. The deviation process $f_{\delta,i}$ represents a client-specific systematic departure from this shared structure, but it is constrained to use the same kernel family and inducing basis as $f_g$ through $k_\delta=\phi k_g$. The parameter $\phi$ is a globally shared hyperparameter. It serves as a shared prior scale for the expected magnitude of systematic deviations across clients. Although $\phi$ is shared, the realized deviation function $f_{\delta,i}$ is client-specific because each client has its own variational posterior. It therefore captures smooth calibration shifts or persistent site-specific offsets that remain aligned with the shared functional structure. By contrast, the local process $f_i$ uses a client-specific kernel $k_i$ and inducing locations $\mathbf{Z}_i$, allowing it to absorb idiosyncratic residual behavior, local disturbances, or shorter-scale variation that should not be forced into the shared global basis. Thus, $f_{\delta,i}$ handles structured client-level calibration relative to the global component, whereas $f_i$ provides flexible local residual adaptation.

\paragraph{Rationale for hierarchical priors.}
The scaled deviation kernel $k_\delta=\phi k_g$ follows the model-discrepancy paradigm of \citet{kennedy2001bayesian}: systematic client shifts are expressed on the same functional basis as the shared component, with population-level intensity $\phi$ learned globally (Section~\ref{sec:algorithm}). Sharing inducing locations $\mathbf{Z}_g$ between $f_g$ and $f_{\delta,i}$ keeps both layers in a common subspace so the asymmetric KL penalties in the ELBO apply to the same basis coefficients. Independent local kernels $k_i$ and inducing sets $\mathbf{Z}_i$ absorb idiosyncratic residuals that need not follow the global geometry. Section~\ref{sec:scenarios_methods} (Scenario~B) stress-tests deviation-kernel mismatch when the fitted model retains an RBF $k_g$ but the simulator uses a periodic deviation kernel.

The shared kernel form $k_\delta=\phi k_g$ is a regularizing design choice rather than an assertion that all clients deviate in exactly the same way. It constrains systematic deviations to the same functional basis as the global component, which makes the deviation interpretable as a calibration offset around the shared structure. Using separate inducing locations or a completely independent deviation kernel would increase flexibility, but it would also weaken the competition between the global and deviation layers and increase communication and optimization complexity. We therefore use the shared basis for the structured deviation layer and reserve the independent kernel $k_i$ and inducing set $\mathbf{Z}_i$ for idiosyncratic local residuals. Section~\ref{sec:scenarios_methods} evaluates this choice under mismatch scenarios in which the true deviation kernel, additivity, or component independence is violated.

\paragraph{Identifiability and regularized attribution.}
Because $k_\delta=\phi k_g$, the marginal prior for a single client's sum $f_g+f_{\delta,i}$ has covariance $(1+\phi)k_g$. Hence, if a client were analyzed in isolation and only this summed process were observed, the split between $f_g$ and $f_{\delta,i}$ would not be uniquely identifiable. This gap is about attribution, not about whether the client's observable function can be predicted. Squared-exponential kernels are universal on compact input domains, so an RBF hierarchy can still approximate \(f_g+f_{\delta,i}\) as a single continuous function of \(x\). What a single client cannot do is uniquely name the two layers inside that sum. The separation in pFedHGP instead relies on the multi-client federated hierarchy: $f_g$ is one shared process used by all clients, whereas $f_{\delta,i}$ is independent across clients. Therefore, for $i\ne j$,
\[
\operatorname{Cov}\{y_i(x),y_j(x')\}=k_g(x,x'),
\]
while the deviation process contributes only to within-client covariance. Under this generative specification, off-diagonal cross-client covariance therefore depends only on the shared global process $f_g$ (not on a claim that $f_g$ is uniquely identified from data). The ELBO further regularizes this attribution through an asymmetric complexity cost: a common pattern represented by $f_g$ incurs one global KL penalty, whereas representing the same pattern redundantly through the client-specific deviations incurs $\sum_i \mathrm{KL}\bigl(q(\boldsymbol{\delta}_i)\Vert p(\boldsymbol{\delta}_i)\bigr)$. This encourages functional patterns common across clients to be represented by $f_g$ and nonredundant client-specific departures to be represented by $f_{\delta,i}$. We therefore interpret the decomposition as a regularized hierarchical attribution supported by multi-client data, rather than as an unconstrained unique decomposition from a single-client marginal process.

For client $i$ with inputs $\mathbf{X}_i = [\mathbf{x}_{i,1},\dots,\mathbf{x}_{i,n_i}]^\top$, we define the stacked latent vectors $\mathbf{f}_{g,i} = f_g(\mathbf{X}_i)$, $\mathbf{f}_{\delta,i} = f_{\delta,i}(\mathbf{X}_i)$, and $\mathbf{f}_i = f_i(\mathbf{X}_i)$. The total latent effect is denoted by $\mathbf{f}_i^{\text{tot}} = \mathbf{f}_{g,i} + \mathbf{f}_{\delta,i} + \mathbf{f}_i$, with observation vector $\mathbf{y}_i \mid \mathbf{f}_i^{\text{tot}} \sim \mathcal{N}(\mathbf{f}_i^{\text{tot}},\sigma^2\mathbf{I}_{n_i})$.

\subsection{Low-Rank Inducing Variable Representation}
\label{sec:inducing_variables}

To obtain a scalable inference scheme, we augment this hierarchical model with inducing variables for each GP component, yielding a low-rank basis decomposition analogous to personalized dimensionality reduction methods.

\begin{proposition}[Low-Rank Decomposition for Mean Representation]
\label{prop:low_rank}
Let $\mathbf{Z}_g \in \mathbb{R}^{M\times d}$ denote shared global inducing locations and $\mathbf{Z}_i \in \mathbb{R}^{M_i\times d}$ denote client-specific inducing locations. Define the inducing variables
\begin{equation}
\mathbf{u}_g = f_g(\mathbf{Z}_g) \in \mathbb{R}^M,\quad
\boldsymbol{\delta}_i = f_{\delta,i}(\mathbf{Z}_g) \in \mathbb{R}^M,\quad
\mathbf{u}_i = f_i(\mathbf{Z}_i) \in \mathbb{R}^{M_i}.
\end{equation}
Under the GP priors specified in Section~\ref{sec:motivation_and_formulation}, these inducing variables are jointly Gaussian:
\begin{equation}
\mathbf{u}_g \sim \mathcal{N}(\mathbf{0}, \mathbf{K}_{gg}),\quad
\boldsymbol{\delta}_i \sim \mathcal{N}(\mathbf{0}, \phi \mathbf{K}_{gg}),\quad
\mathbf{u}_i \sim \mathcal{N}(\mathbf{0}, \mathbf{K}_{ii}),
\end{equation}
where $\mathbf{K}_{gg} = k_g(\mathbf{Z}_g,\mathbf{Z}_g)$ and $\mathbf{K}_{ii} = k_i(\mathbf{Z}_i,\mathbf{Z}_i)$. 

Conditioned on the inducing variables, the latent function values at client $i$'s observations $\mathbf{X}_i$ have the following structure. Let $\mathbf{f}_{g,i} = f_g(\mathbf{X}_i)$, $\mathbf{f}_{\delta,i} = f_{\delta,i}(\mathbf{X}_i)$, and $\mathbf{f}_i = f_i(\mathbf{X}_i)$ denote the latent function vectors evaluated at $\mathbf{X}_i$. Define the interpolation matrices
\begin{equation}
\mathbf{A}_{g,i} = \mathbf{K}_g(\mathbf{X}_i,\mathbf{Z}_g)\mathbf{K}_{gg}^{-1}, \quad
\mathbf{A}_{\delta,i} = \mathbf{K}_g(\mathbf{X}_i,\mathbf{Z}_g)\mathbf{K}_{gg}^{-1}, \quad
\mathbf{A}_{i} = \mathbf{K}_i(\mathbf{X}_i,\mathbf{Z}_i)\mathbf{K}_{ii}^{-1}.
\end{equation}
The conditional mean of the total latent function admits the low-rank basis expansion
\begin{equation}
\label{eq:low_rank_expansion}
\mathbb{E}\big[\mathbf{f}_i^{\mathrm{tot}}\mid \mathbf{u}_g,\boldsymbol{\delta}_i,\mathbf{u}_i\big]
= \underbrace{\mathbf{A}_{g,i}}_{\text{shared basis}} (\mathbf{u}_g + \boldsymbol{\delta}_i)
+ \underbrace{\mathbf{A}_{i}}_{\text{local basis}} \mathbf{u}_i,
\end{equation}
where $\mathbf{f}_i^{\mathrm{tot}} = \mathbf{f}_{g,i} + \mathbf{f}_{\delta,i} + \mathbf{f}_i$ is the total latent function. The proof follows from standard GP conditional mean formulas; see Appendix~\ref{app:proof_prop2} for details.
\end{proposition}

\paragraph{Connection to personalized PCA.}
The structure in~\eqref{eq:low_rank_expansion} parallels personalized PCA (perPCA) \citep{shi2024personalized}, which decomposes client data into shared and unique features using linear bases. In perPCA, each client's data matrix is represented as $\mathbf{X}_i = \mathbf{U}\mathbf{V}_g\T + \mathbf{U}_i\mathbf{V}_i\T + \mathbf{E}_i$, where $\mathbf{U}\mathbf{V}_g\T$ is a shared low-rank component and $\mathbf{U}_i\mathbf{V}_i\T$ is a client-specific low-rank component. Similarly, pFedHGP decomposes the latent function space into:
\begin{itemize}

\item A \textbf{shared low-rank subspace} spanned by the global kernel basis $\mathbf{k}_g(\mathbf{x},\mathbf{Z}_g)\T\mathbf{K}_{gg}^{-1}$ with two components: the fixed global weights $\mathbf{u}_g$ (common to all clients) and client-specific deviation weights $\boldsymbol{\delta}_i$ (capturing systematic offsets).

\item A \textbf{client-specific low-rank subspace} spanned by the local kernel basis $\mathbf{k}_i(\mathbf{x},\mathbf{Z}_i)\T\mathbf{K}_{ii}^{-1}$ with local weights $\mathbf{u}_i$.
\end{itemize}

However, pFedHGP extends beyond perPCA in three critical ways: (1) it uses kernel-induced nonlinear bases rather than linear bases, enabling flexible nonparametric function approximation; (2) it provides full probabilistic modeling with uncertainty quantification for all components via GP priors, including structured residual covariance, whereas perPCA yields point estimates with only simple (often i.i.d.) noise; and (3) it naturally handles functional data and regression tasks through the GP framework, while perPCA is designed for matrix factorization. The explicit regularization $\boldsymbol{\delta}_i \sim \mathcal{N}(\mathbf{0}, \phi \mathbf{K}_{gg})$ ensures that client-specific adjustments remain structurally consistent with the shared physical model encoded in the global kernel.

\subsection{Variational Sparse GP Framework}
\label{sec:variational_sparse_assumption}

To enable scalable and tractable inference in the hierarchical model, we adopt the variational sparse Gaussian process framework based on the Variational Free Energy (VFE) method \citep{titsias2009variational}. This framework avoids restrictive assumptions about the observation covariance structure (unlike FITC) while maintaining computational efficiency.

\begin{assumption}[Mean Field Variational Inference]
\label{assump:mean_field}
We assume a factorized variational posterior distribution over the inducing variables:
\begin{equation}
q(\mathbf{u}_g, \{\boldsymbol{\delta}_i\}, \{\mathbf{u}_i\}) = q(\mathbf{u}_g)\prod_{i=1}^T q(\boldsymbol{\delta}_i)q(\mathbf{u}_i),
\end{equation}
where each factor is a multivariate Gaussian distribution.
\end{assumption}

This mean-field assumption decouples the inference across the hierarchical components while preserving the full covariance structure of the inducing variables within each component. Crucially, we do \emph{not} assume that the conditional posterior over observations is diagonal (the FITC assumption). Instead, we leverage the property that the variational lower bound (ELBO) only depends on the marginal variances of the residual process.

\begin{proposition}[Low-Rank Covariance Decomposition]
\label{prop:covariance}
Due to the independence of the hierarchical components ($f_g \perp f_{\delta,i} \perp f_i$) in the prior, the marginal variance of the total latent function at any input $\mathbf{x}$ is the sum of the component variances. Within the VFE framework \citep{titsias2009variational}, the expected log-likelihood term in the ELBO relies only on these marginal variances, allowing the residual term $\operatorname{Tr}(\mathbf{R}_i)$ to be computed additively as:
\begin{equation}
\label{eq:covariance_decomposition}
\operatorname{Tr}(\mathbf{R}_i) = \operatorname{Tr}(\mathbf{K}_{g,ii} - \mathbf{Q}_{g,ii}) 
+ \operatorname{Tr}(\mathbf{K}_{\delta,ii} - \mathbf{Q}_{\delta,ii}) 
+ \operatorname{Tr}(\mathbf{K}_{i,ii} - \mathbf{Q}_{i,ii}),
\end{equation}
where $\mathbf{R}_i$ represents the dense residual covariance matrix of the total latent function. The low-rank approximation matrices $\mathbf{Q}_{\cdot,ii}$ are defined as in Proposition~\ref{prop:low_rank}.
\end{proposition}

See Appendix~\ref{app:proof_prop3} for the derivation showing that the expected log-likelihood term in the ELBO involves the trace of the residual covariance $\mathbb{E}_q[\log p(\mathbf{y}|\mathbf{f})] \propto -\frac{1}{2\sigma^2}\operatorname{Tr}(\mathbf{K}_{nn} - \mathbf{Q}_{nn})$, which decomposes additively due to the linearity of the trace operator and the independence of components.

\begin{figure}[ht]
\centering
\begin{tikzpicture}[
    matrix/.style={draw, rectangle, minimum size=1.5cm, align=center},
    operator/.style={circle, draw, inner sep=0pt, minimum size=0.5cm}
]
\node[matrix, fill=blue!10] (Kg) {$\mathbf{K}_g$};
\node[above] at (Kg.north) {Global Physics};

\node[operator, right=0.5cm of Kg] (p1) {$+$};

\node[matrix, fill=orange!10, right=0.5cm of p1] (Kd) {$\mathbf{K}_{\delta}$};
\node[above] at (Kd.north) {Calibration};

\node[operator, right=0.5cm of Kd] (p2) {$+$};

\node[matrix, fill=green!10, right=0.5cm of p2] (Ki) {$\mathbf{K}_i$};
\node[above] at (Ki.north) {Local Noise};

\node[operator, right=0.5cm of Ki] (eq) {$=$};

\node[matrix, fill=gray!20, right=0.5cm of eq, minimum size=2cm] (Ktot) {$\mathbf{K}_{total}$};
\node[above] at (Ktot.north) {Observation Covariance};

\node[below=0.2cm of Kg, font=\scriptsize] {Low-Rank $\mathbf{Q}_g$};
\node[below=0.2cm of Ki, font=\scriptsize] {Low-Rank $\mathbf{Q}_i$};
\node[below=0.2cm of Ktot, font=\scriptsize, align=center] {Dense Matrix\\(Approximated via VFE)};

\draw[->, thick] (Kg) -- (p1);
\draw[->, thick] (p1) -- (Kd);
\draw[->, thick] (Kd) -- (p2);
\draw[->, thick] (p2) -- (Ki);
\draw[->, thick] (Ki) -- (eq);
\draw[->, thick] (eq) -- (Ktot);
\end{tikzpicture}
\caption{Structure of the additive covariance in pFedHGP. While the components are independent (block-diagonal in latent space), they sum to form the observation covariance. The ELBO in the VFE framework allows us to efficiently compute this by utilizing only the diagonal elements of the residual $\mathbf{K}-\mathbf{Q}$.}
\label{fig:covariance_structure}
\end{figure}
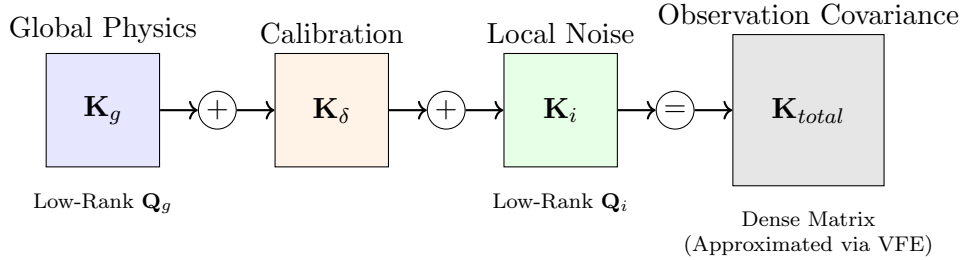

This result is significant because it justifies the efficiency of pFedHGP without making the restrictive FITC assumption that off-diagonal data correlations are zero. We simply exploit the property that maximizing the ELBO only requires matching the marginal variances of the residual process. Figure~\ref{fig:covariance_structure} illustrates this additive structure.

\subsection{Federated Variational Inference}
\label{sec:variational_inference}

Given the hierarchical GP generative model in~\eqref{eq:obs_decomposition}, the joint distribution over all latent variables and observations can be written as
\begin{equation}
\begin{aligned}
p\big(\{\mathbf{y}_i\}, f_g, \{f_{\delta,i}\}, \{f_i\}, \mathbf{u}_g, \{\boldsymbol{\delta}_i\}, \{\mathbf{u}_i\}\big)
&= p(\mathbf{u}_g)\prod_{i=1}^T p(\boldsymbol{\delta}_i)\,p(\mathbf{u}_i)\; \\
&\quad \times p(f_g\mid \mathbf{u}_g)\prod_{i=1}^T p(f_{\delta,i}\mid \boldsymbol{\delta}_i)\,p(f_i\mid \mathbf{u}_i)\; \\
&\quad \times \prod_{i=1}^T p(\mathbf{y}_i\mid \mathbf{f}_i^{\mathrm{tot}}),
\end{aligned}
\end{equation}
where $\mathbf{f}_i^{\mathrm{tot}} = f_g(\mathbf{X}_i) + f_{\delta,i}(\mathbf{X}_i) + f_i(\mathbf{X}_i)$ and $p(\mathbf{y}_i\mid \mathbf{f}_i^{\mathrm{tot}}) = \mathcal{N}(\mathbf{f}_i^{\mathrm{tot}},\sigma^2\mathbf{I})$.

Exact posterior inference is analytically tractable in principle, but computationally prohibitive in the federated, large-scale setting due to the $O(N^3)$ coupling across all clients and components. We therefore follow the variational sparse GP strategy and introduce a variational distribution of the form
\begin{equation}
q(f_g,\{f_{\delta,i}\},\{f_i\},\mathbf{u}_g,\{\boldsymbol{\delta}_i\},\{\mathbf{u}_i\})
= p(f_g\mid \mathbf{u}_g)\prod_{i=1}^T p(f_{\delta,i}\mid \boldsymbol{\delta}_i)\,p(f_i\mid \mathbf{u}_i)\;
  q(\mathbf{u}_g)\prod_{i=1}^T q(\boldsymbol{\delta}_i)\,q(\mathbf{u}_i),
\end{equation}
where
\[
q(\mathbf{u}_g) = \mathcal{N}(\mathbf{m}_g, \mathbf{S}_g), \quad
q(\boldsymbol{\delta}_i) = \mathcal{N}(\mathbf{m}_{\delta_i}, \mathbf{S}_{\delta_i}), \quad
q(\mathbf{u}_i) = \mathcal{N}(\mathbf{m}_{u_i}, \mathbf{S}_{u_i})
\]
are free Gaussian factors. The Gaussian conditionals $p(f_g\mid \mathbf{u}_g)$, $p(f_{\delta,i}\mid \boldsymbol{\delta}_i)$, and $p(f_i\mid \mathbf{u}_i)$ are those implied by the GP priors and inducing locations, and the latent function values are analytically integrated out.

\paragraph{Hierarchical ELBO.}

Following Proposition~\ref{prop:vfe_elbo}, we integrate out the latent function values $\mathbf{f}_i^{\mathrm{tot}}$ at $\mathbf{X}_i$ conditional on $(\mathbf{u}_g,\boldsymbol{\delta}_i,\mathbf{u}_i)$ and adopt the VFE variational distribution $q(\mathbf{f}_i^{\mathrm{tot}}\mid \mathbf{u}_g,\boldsymbol{\delta}_i,\mathbf{u}_i)\,q(\mathbf{u}_g)q(\boldsymbol{\delta}_i)q(\mathbf{u}_i)$ with mean-field $q$ factors. The inducing-conditional model for client $i$ is then

\begin{equation}
\label{eq:hierarchical_likelihood}
p(\mathbf{y}_i \mid \mathbf{u}_g,\boldsymbol{\delta}_i,\mathbf{u}_i)
:= \mathcal{N}\big(\mathbf{A}_{g,i}(\mathbf{u}_g+\boldsymbol{\delta}_i) + \mathbf{A}_{i}\mathbf{u}_i, \,\sigma^2\mathbf{I}_{n_i}\big),
\end{equation}

which is the hierarchical analogue of $p(\mathbf{y}\mid\mathbf{u})$ in~\eqref{eq:sgp_vfe} (a low-rank Gaussian in the inducing variables only). It does \emph{not} yet include the sparse GP residual covariance: under VFE, marginalizing $\mathbf{f}_i^{\mathrm{tot}}$ contributes the additive penalty $-\frac{1}{2\sigma^2}\operatorname{Tr}(\mathbf{R}_i)$ with $\operatorname{Tr}(\mathbf{R}_i)$ given by~\eqref{eq:covariance_decomposition}. Moreover, because $p(\mathbf{y}_i \mid \mathbf{u}_g,\boldsymbol{\delta}_i,\mathbf{u}_i)$ is linear in $(\mathbf{u}_g,\boldsymbol{\delta}_i,\mathbf{u}_i)$, the expectation $\mathbb{E}_q[\log p(\mathbf{y}_i \mid \mathbf{u}_g,\boldsymbol{\delta}_i,\mathbf{u}_i)]$ also propagates posterior covariance in each inducing block. Proposition~\ref{prop:closed_form_ll} states the resulting closed form (squared-error, inducing-covariance traces, and $\operatorname{Tr}(\mathbf{R}_i)$); Theorem~\ref{prop:elbo_decomp} combines this with the KL decomposition below.

The evidence lower bound (ELBO) for pFedHGP, expressed in terms of the global, client-specific deviation, and local inducing variables, is

\begin{equation}
\label{eq:hierarchical_elbo}
\begin{aligned}
\mathcal{L}
&:= \sum_{i=1}^T \mathbb{E}_{q(\mathbf{u}_g)\,q(\boldsymbol{\delta}_i)\,q(\mathbf{u}_i)}\bigl[\log p(\mathbf{y}_i \mid \mathbf{u}_g,\boldsymbol{\delta}_i,\mathbf{u}_i)\bigr] \\
&\quad - \mathrm{KL}\bigl(q(\mathbf{u}_g,\{\boldsymbol{\delta}_i\},\{\mathbf{u}_i\}) \,\|\, p(\mathbf{u}_g)\prod_{i=1}^T p(\boldsymbol{\delta}_i)p(\mathbf{u}_i)\bigr),
\end{aligned}
\end{equation}

where the expectation $\mathbb{E}_q[\log p(\mathbf{y}_i \mid \mathbf{u}_g,\boldsymbol{\delta}_i,\mathbf{u}_i)]$ is understood as the full VFE expected log-likelihood in~\eqref{eq:closed_form_ll} (not the plug-in log-density of~\eqref{eq:hierarchical_likelihood} evaluated only at posterior means). Proposition~\ref{prop:kl_decomp} shows that the overall KL term decomposes into component-wise KL divergences, enabling efficient federated optimization.

\begin{proposition}[Decomposability of KL Terms]
\label{prop:kl_decomp}
Under the mean-field variational posterior (Assumption~\ref{assump:mean_field}), the KL divergence between the variational posterior and the prior factorizes across components:
\begin{equation}
\label{eq:kl_decomposition}
\begin{aligned}
&\mathrm{KL}\bigl(q(\mathbf{u}_g,\{\boldsymbol{\delta}_i\},\{\mathbf{u}_i\}) \,\|\, p(\mathbf{u}_g)\prod_{i=1}^T p(\boldsymbol{\delta}_i)p(\mathbf{u}_i)\bigr)
= \\
&\quad \mathrm{KL}\bigl(q(\mathbf{u}_g)\|p(\mathbf{u}_g)\bigr)
  + \sum_{i=1}^T \mathrm{KL}\bigl(q(\boldsymbol{\delta}_i)\|p(\boldsymbol{\delta}_i)\bigr)
  + \sum_{i=1}^T \mathrm{KL}\bigl(q(\mathbf{u}_i)\|p(\mathbf{u}_i)\bigr).
\end{aligned}
\end{equation}
\end{proposition}

\begin{proposition}[Decomposability of Expected Log-Likelihood]
\label{prop:closed_form_ll}

Under the factorized variational posterior (Assumption~\ref{assump:mean_field}), the VFE marginalization of $\mathbf{f}_i^{\mathrm{tot}}$, and the additivity of marginal variances (Proposition~\ref{prop:covariance}), the expected log-likelihood term in the ELBO~\eqref{eq:hierarchical_elbo} admits a decomposable closed-form expression. Equation~\eqref{eq:hierarchical_likelihood} supplies the inducing-conditional mean model; the expectation below adds posterior uncertainty in the inducing blocks and the sparse GP residual traces:

Define the interpolation matrices $\mathbf{A}_{g,i}$, $\mathbf{A}_{\delta,i}$ (since $k_\delta=\phi k_g$, $\mathbf{A}_{\delta,i}=\mathbf{A}_{g,i}$), and $\mathbf{A}_{i}$ from Proposition~\ref{prop:low_rank}. Then
\begin{equation}
\label{eq:closed_form_ll}
\begin{aligned}
\mathbb{E}_{q}\bigl[\log p(\mathbf{y}_i \mid \mathbf{u}_g,\boldsymbol{\delta}_i,\mathbf{u}_i)\bigr]
&= -\frac{n_i}{2}\log(2\pi \sigma^2) - \frac{1}{2\sigma^2}\|\mathbf{y}_i - \boldsymbol{\mu}_{\mathbf{f}_i}\|^2 \\
&\quad - \frac{1}{2\sigma^2}\Bigl(
\operatorname{Tr}\bigl(\mathbf{A}_{g,i}\mathbf{S}_g\mathbf{A}_{g,i}^\top\bigr)
+ \operatorname{Tr}\bigl(\mathbf{A}_{\delta,i}\mathbf{S}_{\delta_i}\mathbf{A}_{\delta,i}^\top\bigr)
+ \operatorname{Tr}\bigl(\mathbf{A}_{i}\mathbf{S}_{u_i}\mathbf{A}_{i}^\top\bigr)\Bigr) \\
&\quad - \frac{1}{2\sigma^2}\operatorname{Tr}\bigl(\mathbf{R}_i\bigr),
\end{aligned}
\end{equation}%

where the variational predictive mean decomposes as
\begin{equation}
\label{eq:mean_decomposition}
\boldsymbol{\mu}_{\mathbf{f}_i}
= \underbrace{\mathbf{A}_{g,i}\mathbf{m}_g}_{\mathrm{global}}
+ \underbrace{\mathbf{A}_{\delta,i}\mathbf{m}_{\delta_i}}_{\mathrm{client}\ \mathrm{deviation}}
+ \underbrace{\mathbf{A}_{i}\mathbf{m}_{u_i}}_{\mathrm{local}\ \mathrm{residual}},
\end{equation}
and the residual trace decomposes additively via Proposition~\ref{prop:covariance}:
\begin{equation}
\label{eq:trace_decomposition}
\operatorname{Tr}\big(\mathbf{R}_i\big) = \operatorname{Tr}(\mathbf{K}_{g,ii} - \mathbf{Q}_{g,ii}) 
+ \operatorname{Tr}(\mathbf{K}_{\delta,ii} - \mathbf{Q}_{\delta,ii}) 
+ \operatorname{Tr}(\mathbf{K}_{i,ii} - \mathbf{Q}_{i,ii}).
\end{equation}
\end{proposition}

\begin{theorem}[Explicit ELBO Decomposition]
\label{prop:elbo_decomp}
Under the mean-field assumption (Assumption~\ref{assump:mean_field}), substitute the closed-form expected log-likelihood from Proposition~\ref{prop:closed_form_ll} into the ELBO~\eqref{eq:hierarchical_elbo}. The bound admits the following explicit decomposed form, combining the decompositions from Propositions~\ref{prop:kl_decomp} and~\ref{prop:closed_form_ll}:

\begin{equation}
\label{eq:elbo_explicit_decomp}
\begin{aligned}
\mathcal{L} = {} & \sum_{i=1}^T \Bigl[-\tfrac{n_i}{2}\log(2\pi \sigma^2) - \tfrac{1}{2\sigma^2}\|\mathbf{y}_i - \boldsymbol{\mu}_{\mathbf{f}_i}\|^2 \\
&\qquad - \tfrac{1}{2\sigma^2} \bigl(
\operatorname{Tr}(\mathbf{A}_{g,i}\mathbf{S}_g\mathbf{A}_{g,i}^\top)
+ \operatorname{Tr}(\mathbf{A}_{\delta,i}\mathbf{S}_{\delta_i}\mathbf{A}_{\delta,i}^\top)
+ \operatorname{Tr}(\mathbf{A}_{i}\mathbf{S}_{u_i}\mathbf{A}_{i}^\top) \\
&\qquad\qquad + \operatorname{Tr}(\mathbf{K}_{g,ii} - \mathbf{Q}_{g,ii}) 
 + \operatorname{Tr}(\mathbf{K}_{\delta,ii} - \mathbf{Q}_{\delta,ii}) 
 + \operatorname{Tr}(\mathbf{K}_{i,ii} - \mathbf{Q}_{i,ii})\bigr)\Bigr] \\
&\quad - \mathrm{KL}\bigl(q(\mathbf{u}_g)\|p(\mathbf{u}_g)\bigr)
  - \sum_{i=1}^T \mathrm{KL}\bigl(q(\boldsymbol{\delta}_i)\|p(\boldsymbol{\delta}_i)\bigr)
  - \sum_{i=1}^T \mathrm{KL}\bigl(q(\mathbf{u}_i)\|p(\mathbf{u}_i)\bigr),
\end{aligned}
\end{equation}%

where $\boldsymbol{\mu}_{\mathbf{f}_i}$ is the decomposed variational predictive mean defined in~\eqref{eq:mean_decomposition}.
\end{theorem}

This result explicitly connects the variational objective to the efficient trace computation derived in Proposition~\ref{prop:covariance}. The least squares term fits the decomposed variational predictive mean, the $\mathbf{A}_{\cdot,i}\mathbf{S}_{\cdot}\mathbf{A}_{\cdot,i}^\top$ traces propagate posterior uncertainty in each inducing block, and the residual traces penalize sparse-approximation uncertainty not captured by inducing variables. Optimization still requires only diagonal elements of $\mathbf{R}_i$ for the residual terms, preserving the efficiency of our framework.

\paragraph{Summary and intuition (ELBO).}
The explicit ELBO in Theorem~\ref{prop:elbo_decomp} is the VFE bound in~\eqref{eq:hierarchical_elbo} with the expected log-likelihood expanded as in Proposition~\ref{prop:closed_form_ll}: a squared-error term on the hierarchical predictive mean, inducing-posterior covariance traces, additive sparse GP residual traces $\operatorname{Tr}(\mathbf{R}_i)$, and KL penalties (one global term on $\mathbf{u}_g$ versus $T$ client terms on $\{\boldsymbol{\delta}_i\}$), encouraging population-level structure to reside in $f_g$.

A detailed derivation is provided in Appendix~\ref{app:elbo_proof}.

\paragraph{Operational identifiability.}
Theorem~\ref{prop:elbo_decomp} should be read as defining an \emph{ELBO-regularized allocation} of variation across layers rather than a proof of classical point identification of $(f_g,f_{\delta,i})$ from a single client's marginal law. When $k_\delta=\phi k_g$, the sum $f_g+f_{\delta,i}$ is itself a GP marginally; cross-client sharing, distinct local kernels $k_i$, and asymmetric KL penalties on $(\mathbf{u}_g,\{\boldsymbol{\delta}_i\})$ nonetheless provide a practical preference for placing multi-client structure in $f_g$ while reserving $f_{\delta,i}$ for stable systematic shifts, as detailed in the paragraph \emph{Identifiability and regularized attribution} in Section~\ref{sec:motivation_and_formulation}.

\subsection{pFedHGP Algorithm} 
\label{sec:algorithm} 
Training proceeds in synchronous communication rounds indexed by \(r=0,1,2,\dots\). At the start of round \(r\), the server holds the current global parameter block 
\[
\Theta_{\mathrm{global}}^{(r)} 
= \bigl(\mathbf{m}_g^{(r)}, \mathbf{S}_g^{(r)}, \ell_g^{(r)}, \sigma_g^{2(r)}, \eta_\phi^{(r)}, \mathbf{Z}_g^{(r)}\bigr),
\] 
comprising both the global variational parameters \((\mathbf{m}_g^{(r)}, \mathbf{S}_g^{(r)})\) and the global hyperparameters \((\ell_g^{(r)}, \sigma_g^{2(r)}, \eta_\phi^{(r)}, \mathbf{Z}_g^{(r)})\), with deviation scale \(\phi^{(r)}=\operatorname{softplus}(\eta_\phi^{(r)})\). Each client \(i\) holds its local parameter block 
\[
\Theta_{i,\mathrm{local}}^{(r)} 
= \bigl(\mathbf{m}_{\delta_i}^{(r)}, \mathbf{S}_{\delta_i}^{(r)}, \mathbf{m}_{u_i}^{(r)}, \mathbf{S}_{u_i}^{(r)}, \ell_i^{(r)}, \sigma_i^{2(r)}, \mathbf{Z}_i^{(r)}\bigr),
\] 
containing local variational parameters \((\mathbf{m}_{\delta_i}^{(r)}, \mathbf{S}_{\delta_i}^{(r)}, \mathbf{m}_{u_i}^{(r)}, \mathbf{S}_{u_i}^{(r)})\) and local hyperparameters \((\ell_i^{(r)}, \sigma_i^{2(r)}, \mathbf{Z}_i^{(r)})\). The algorithm alternates between client-side local optimization and server-side global aggregation; broadcasting is integrated at the end of the server update. 
\subsubsection{Local Optimization} 
When client \(i\) receives \(\Theta_{\mathrm{global}}^{(r)}\) from the server at round \(r\), it treats these global parameters as fixed and optimizes its own local block using its private data \(\{\mathbf{X}_i, \mathbf{y}_i\}\). For convenience we collect all client-specific variables into 
\[
\Theta_{i,\mathrm{local}}^{(r)} 
= \bigl(\mathbf{m}_{\delta_i}^{(r)}, \mathbf{S}_{\delta_i}^{(r)},
        \mathbf{m}_{u_i}^{(r)}, \mathbf{S}_{u_i}^{(r)},
        \ell_i^{(r)}, \sigma_i^{2(r)}, \mathbf{Z}_i^{(r)}\bigr),
\]
where \(q(\boldsymbol{\delta}_i) = \mathcal{N}(\mathbf{m}_{\delta_i}, \mathbf{S}_{\delta_i})\) and \(q(\mathbf{u}_i) = \mathcal{N}(\mathbf{m}_{u_i}, \mathbf{S}_{u_i})\) are the local variational distributions, and \((\ell_i, \sigma_i^{2}, \mathbf{Z}_i)\) are the kernel hyperparameters and inducing inputs of the idiosyncratic local kernel. The prior \(p(\boldsymbol{\delta}_i)\) is specified by the deviation prior with hyperparameter \(\phi^{(r)}=\operatorname{softplus}(\eta_\phi^{(r)})\), and \(p(\mathbf{u}_i)\) by the local kernel with parameters \((\ell_i, \sigma_i^{2}, \mathbf{Z}_i)\). In implementation each covariance matrix is represented by its Cholesky factor \(\mathbf{S}_{(\cdot)} = \mathbf{L}_{(\cdot)}\mathbf{L}_{(\cdot)}^\top\); the optimization is carried out with respect to the entries of these Cholesky factors, although for notational simplicity we keep the covariance notation \(\mathbf{S}_{(\cdot)}\).

\begin{proposition}[Local ELBO for Client Optimization]
\label{prop:local_elbo}
Based on the explicit ELBO decomposition in Theorem~\ref{prop:elbo_decomp}, client \(i\) maximizes the local ELBO, which is a function of the local parameters given the fixed global parameters:

\begin{equation} 
\label{eq:local_elbo_new}
\begin{aligned}
\mathcal{L}_i\bigl(\Theta_{i,\mathrm{local}} \mid \Theta_{\mathrm{global}}^{(r)}\bigr) 
&= - \tfrac{1}{2\sigma^2}\bigl\|\mathbf{y}_i - \boldsymbol{\mu}_{\mathbf{f}_i}\bigr\|^2 \\
&\quad - \tfrac{1}{2\sigma^2}\Bigl(
     \operatorname{Tr}(\mathbf{A}_{\delta,i}\mathbf{S}_{\delta_i}\mathbf{A}_{\delta,i}^\top)
     + \operatorname{Tr}(\mathbf{A}_{i}\mathbf{S}_{u_i}\mathbf{A}_{i}^\top)
     + \operatorname{Tr}(\mathbf{K}_{i,ii} - \mathbf{Q}_{i,ii})\Bigr) \\
&\quad - \mathrm{KL}\bigl(q(\boldsymbol{\delta}_i)\,\|\,p(\boldsymbol{\delta}_i)\bigr)
      - \mathrm{KL}\bigl(q(\mathbf{u}_i)\,\|\,p(\mathbf{u}_i)\bigr),
\end{aligned}
\end{equation}%

where the decomposed variational predictive mean \(\boldsymbol{\mu}_{\mathbf{f}_i}\) is defined in~\eqref{eq:mean_decomposition} (with $\mathbf{m}_g$ replaced by $\mathbf{m}_g^{(r)}$ which is fixed from $\Theta_{\mathrm{global}}^{(r)}$), and the low-rank approximation matrix \(\mathbf{Q}_{i,ii} = \mathbf{K}_i(\mathbf{X}_i, \mathbf{Z}_i) \mathbf{K}_{ii}^{-1} \mathbf{K}_i(\mathbf{Z}_i, \mathbf{X}_i)\) depends on local kernel hyperparameters $(\ell_i, \sigma_i^2)$ and local inducing locations $\mathbf{Z}_i$. Constant terms with respect to local parameters (including $\operatorname{Tr}(\mathbf{A}_{g,i}\mathbf{S}_g^{(r)}\mathbf{A}_{g,i}^\top)$, $\operatorname{Tr}(\mathbf{K}_{g,ii} - \mathbf{Q}_{g,ii})$, and $\operatorname{Tr}(\mathbf{K}_{\delta,ii} - \mathbf{Q}_{\delta,ii})$, which depend only on $\Theta_{\mathrm{global}}^{(r)}$) have been omitted.
 The notation $\mathcal{L}_i(\Theta_{i,\mathrm{local}} \mid \Theta_{\mathrm{global}}^{(r)})$ emphasizes that the global parameters are given/fixed, and the optimization is performed only over the local parameters $\Theta_{i,\mathrm{local}}$.
\end{proposition} 
The explicit decomposition in~\eqref{eq:local_elbo_new} reveals that the local objective separates into: (1) a data-fitting least squares term $-\frac{1}{2\sigma^2}\|\mathbf{y}_i - \boldsymbol{\mu}_{\mathbf{f}_i}\|^2$ involving the decomposed predictive mean \(\boldsymbol{\mu}_{\mathbf{f}_i}\), (2) a trace regularization term $\operatorname{Tr}(\mathbf{K}_{i,ii} - \mathbf{Q}_{i,ii})$ for the local residual covariance, and (3) KL regularization terms for the client-specific deviation and local components. Since \(\Theta_{\mathrm{global}}^{(r)}\) is fixed, the global contributions $\mathbf{A}_{g,i}\mathbf{m}_g^{(r)}$ in the predictive mean can be computed using only local data and the known global posterior approximation.

Client \(i\) applies a few steps of gradient ascent to \eqref{eq:local_elbo_new} with respect to \(\Theta_{i,\mathrm{local}}\), while holding \(\Theta_{\mathrm{global}}^{(r)}\) fixed. This local optimization simultaneously updates the variational means \(\mathbf{m}_{\delta_i}\) and \(\mathbf{m}_{u_i}\), the Cholesky factors underlying \(\mathbf{S}_{\delta_i}\) and \(\mathbf{S}_{u_i}\), the local kernel hyperparameters \(\ell_i\) (lengthscale / bandwidth) and \(\sigma_i^2\) (output scale), and the local inducing locations \(\mathbf{Z}_i\). We denote the updated local parameters by \(\Theta_{i,\mathrm{local}}^{(r+1)}\).

\subsubsection{Global Component Optimization} 
In this step we optimize the global parameter block 
\[
\Theta_{\mathrm{global}}
= \bigl(\mathbf{m}_g, \mathbf{S}_g,
        \ell_g, \sigma_g^{2}, \eta_\phi, \mathbf{Z}_g\bigr),
\qquad
\phi=\operatorname{softplus}(\eta_\phi),
\]
while keeping all local blocks $\{\Theta_{i,\mathrm{local}}^{(r+1)}\}_{i=1}^T$ fixed at the values obtained from the client-side updates in the current round. We first isolate the part of the ELBO that depends on $\Theta_{\mathrm{global}}$ and then show that its gradient decomposes into client-wise contributions that can be computed locally and aggregated at the server.

\begin{proposition}[Global ELBO given fixed local blocks]
\label{prop:global_elbo_block}
Let $\{\Theta_{i,\mathrm{local}}^{(r+1)}\}_{i=1}^T$ be fixed local parameters after the client-side updates in round $r$, and let $\boldsymbol{\mu}_{\mathbf{f}_i}$ be the decomposed predictive mean in~\eqref{eq:mean_decomposition}. Up to an additive constant independent of $\Theta_{\mathrm{global}}$, the ELBO~\eqref{eq:elbo_explicit_decomp} can be written as

\begin{equation} 
\label{eq:elbo_global_block}
\begin{aligned}
\mathcal{L}\bigl(\Theta_{\mathrm{global}} \mid \{\Theta_{i,\mathrm{local}}^{(r+1)}\}\bigr)
&= \sum_{i=1}^T \mathcal{L}_i^{\mathrm{global}}\bigl(\Theta_{\mathrm{global}} \mid \Theta_{i,\mathrm{local}}^{(r+1)}\bigr) \\
&\quad - \mathrm{KL}\bigl(q(\mathbf{u}_g)\,\|\,p(\mathbf{u}_g)\bigr)
   - \sum_{i=1}^T \mathrm{KL}\bigl(q(\boldsymbol{\delta}_i)\,\|\,p(\boldsymbol{\delta}_i;\phi,\mathbf{K}_{gg})\bigr)
   + \textnormal{const},
\end{aligned}
\end{equation} 
where the client-wise global contributions are
\begin{equation}
\label{eq:Li_global}
\begin{aligned}
\mathcal{L}_i^{\mathrm{global}}\bigl(\Theta_{\mathrm{global}} \mid \Theta_{i,\mathrm{local}}^{(r+1)}\bigr)
&= -\tfrac{1}{2\sigma^2}\bigl\|\mathbf{y}_i - \boldsymbol{\mu}_{\mathbf{f}_i}\bigr\|^2 \\
&\quad - \tfrac{1}{2\sigma^2}
    \Bigl(
    \operatorname{Tr}(\mathbf{A}_{g,i}\mathbf{S}_g\mathbf{A}_{g,i}^\top)
    + \operatorname{Tr}(\mathbf{A}_{\delta,i}\mathbf{S}_{\delta_i}\mathbf{A}_{\delta,i}^\top) \\
&\qquad\qquad + \operatorname{Tr}(\mathbf{K}_{g,ii} - \mathbf{Q}_{g,ii})
         + \operatorname{Tr}(\mathbf{K}_{\delta,ii} - \mathbf{Q}_{\delta,ii})\Bigr).
\end{aligned}
\end{equation}
Here $\mathbf{K}_{g,ii}$, $\mathbf{Q}_{g,ii}$, $\mathbf{K}_{\delta,ii}$, and $\mathbf{Q}_{\delta,ii}$ depend on $\Theta_{\mathrm{global}}$ through the global kernel hyperparameters $(\ell_g,\sigma_g^2)$, $\phi=\operatorname{softplus}(\eta_\phi)$, and inducing locations $\mathbf{Z}_g$; the predictive mean $\boldsymbol{\mu}_{\mathbf{f}_i}$ depends on $\Theta_{\mathrm{global}}$ through $(\mathbf{m}_g,\mathbf{S}_g)$ and the global kernel; and $\mathbf{A}_{g,i}$, $\mathbf{A}_{\delta,i}$ inherit the same dependence through $\mathbf{Z}_g$ and $k_g$. The terms $\operatorname{Tr}(\mathbf{A}_{i}\mathbf{S}_{u_i}\mathbf{A}_{i}^\top)$, $\operatorname{Tr}(\mathbf{K}_{i,ii} - \mathbf{Q}_{i,ii})$, and the local KL on $\mathbf{u}_i$ do not depend on $\Theta_{\mathrm{global}}$ and are absorbed into the constant. The deviation KL terms are retained in~\eqref{eq:elbo_global_block} because $p(\boldsymbol{\delta}_i)$ depends on $(\phi,\mathbf{K}_{gg})$.
 A proof is provided in Appendix~\ref{app:global_elbo_block_proof}.
\end{proposition}

Proposition~\ref{prop:global_elbo_block} shows that optimizing the global block amounts to maximizing a sum of client-wise objectives $\mathcal{L}_i^{\mathrm{global}}$ minus client-specific deviation KL penalties, together with the global inducing KL on $\mathbf{u}_g$. This structure enables a federated scheme in which each client transmits a gradient summary that already includes its private deviation-KL contribution, while the server aggregates these summaries and subtracts only the shared global KL term (which depends on $(\mathbf{m}_g,\mathbf{S}_g)$ held at the server).

For fixed local blocks $\{\Theta_{i,\mathrm{local}}^{(r+1)}\}$, each client $i$ computes and transmits
\begin{equation} 
\label{eq:Si_def}
\widetilde{\mathcal{S}}_i^{(r+1)}
:= \nabla_{\Theta_{\mathrm{global}}}
   \Bigl[
   \mathcal{L}_i^{\mathrm{global}}\bigl(\Theta_{\mathrm{global}} \mid \Theta_{i,\mathrm{local}}^{(r+1)}\bigr)
   - \mathrm{KL}\bigl(q(\boldsymbol{\delta}_i)\,\|\,p(\boldsymbol{\delta}_i;\phi,\mathbf{K}_{gg})\bigr)
   \Bigr]
   \Big|_{\Theta_{\mathrm{global}} = \Theta_{\mathrm{global}}^{(r)}},
\end{equation} 
using its private observations $(\mathbf{X}_i,\mathbf{y}_i)$, updated local block $\Theta_{i,\mathrm{local}}^{(r+1)}$ (which contains $q(\boldsymbol{\delta}_i)$), and broadcast $\Theta_{\mathrm{global}}^{(r)}$. Only the vector $\widetilde{\mathcal{S}}_i^{(r+1)}$ is sent to the server; $\mathbf{m}_{\delta_i}$, $\mathbf{S}_{\delta_i}$, and raw data remain local. The full global gradient is then

\begin{equation} 
\label{eq:global_gradient}
\nabla_{\Theta_{\mathrm{global}}}\mathcal{L}\bigl(\Theta_{\mathrm{global}} \mid \{\Theta_{i,\mathrm{local}}^{(r+1)}\}\bigr)
= \sum_{i=1}^T \widetilde{\mathcal{S}}_i^{(r+1)}
   - \nabla_{\Theta_{\mathrm{global}}}\,
     \mathrm{KL}\bigl(q(\mathbf{u}_g)\,\|\,p(\mathbf{u}_g)\bigr),
\end{equation}%

evaluated at $\Theta_{\mathrm{global}} = \Theta_{\mathrm{global}}^{(r)}$.

Upon collecting $\{\widetilde{\mathcal{S}}_i^{(r+1)}\}_{i=1}^T$, the server aggregates
\[
\widetilde{\mathcal{S}}^{(r+1)} := \sum_{i=1}^T \widetilde{\mathcal{S}}_i^{(r+1)},
\]
forms $\nabla_{\Theta_{\mathrm{global}}}\mathcal{L}$ using~\eqref{eq:global_gradient} (subtracting only the global inducing KL, which is evaluated from $(\mathbf{m}_g,\mathbf{S}_g)$ at the server), and updates the entire global block by a gradient step, e.g.
\begin{equation}
\label{eq:global_update}
\Theta_{\mathrm{global}}^{(r+1)} 
= \Theta_{\mathrm{global}}^{(r)} 
+ \eta_{\mathrm{global}} \,\nabla_{\Theta_{\mathrm{global}}}\mathcal{L},
\end{equation} 
where the gradient in~\eqref{eq:global_update} is taken with respect to the unconstrained global coordinates in $\Theta_{\mathrm{global}}^{(r)}$ (including $\eta_\phi^{(r)}$ rather than $\phi$ directly). Positive hyperparameters such as $\phi$ enter the ELBO only through reparameterizations $\phi=\operatorname{softplus}(\eta_\phi)$ and analogous softplus maps for lengthscales and variances (Section~\ref{sec:implementation}). The updated $\Theta_{\mathrm{global}}^{(r+1)}$ is then broadcast to clients for the next communication round. This procedure shows explicitly that the global component optimization is decomposable across clients and implementable in a federated way via local gradients and server-side aggregation.

\paragraph{Deviation scale $\phi$ as a federated hyperparameter.}
The scale factor $\phi>0$ is a globally shared hyperparameter that controls the prior variance of the systematic deviation layer, $k_\delta=\phi k_g$. It should be interpreted as a population-level regularization scale rather than as a constraint that all clients have identical realized deviation magnitudes. Client-specific amplitudes and shapes are still learned through the local variational posteriors $q(\boldsymbol{\delta}_i)=\mathcal{N}(\mathbf{m}_{\delta_i},\mathbf{S}_{\delta_i})$. To enforce $\phi>0$ throughout optimization, we include an unconstrained coordinate $\eta_\phi\in\mathbb{R}$ in $\Theta_{\mathrm{global}}$ and set $\phi=\operatorname{softplus}(\eta_\phi)$. After the client-side updates in round $r+1$, the server aggregates client-wise contributions and updates
\begin{equation}
\label{eq:eta_phi_update}
\eta_\phi^{(r+1)}
=
\eta_\phi^{(r)}
+
\eta_{\mathrm{global}}
\sum_{i=1}^T
\mathcal{S}_{i,\eta_\phi}^{(r+1)},
\qquad
\phi^{(r+1)}=\operatorname{softplus}\!\bigl(\eta_\phi^{(r+1)}\bigr),
\end{equation}
where client $i$ contributes
\begin{equation}
\label{eq:Si_eta_phi}
\mathcal{S}_{i,\eta_\phi}^{(r+1)}
=
\left.
\frac{\partial \mathcal{L}_i^{\mathrm{global}}}{\partial \eta_\phi}
-\frac{\partial}{\partial \eta_\phi}
\mathrm{KL}\bigl(q(\boldsymbol{\delta}_i)\,\|\,p(\boldsymbol{\delta}_i;\phi,\mathbf{K}_{gg})\bigr)
\right|_{\Theta_{\mathrm{global}}=\Theta_{\mathrm{global}}^{(r)}},
\end{equation}
which is the $\eta_\phi$-component of $\widetilde{\mathcal{S}}_i^{(r+1)}$ in~\eqref{eq:Si_def}. Under the shared-basis parameterization $k_\delta=\phi k_g$, the deviation interpolation matrix is $\mathbf{A}_{\delta,i}=\phi\mathbf{K}_g(\mathbf{X}_i,\mathbf{Z}_g)(\phi\mathbf{K}_{gg})^{-1}=\mathbf{A}_{g,i}$ (Proposition~\ref{prop:low_rank}), so with $q(\boldsymbol{\delta}_i)$ fixed at the global step the deviation contribution $\mathbf{A}_{\delta,i}\mathbf{m}_{\delta_i}$ to $\boldsymbol{\mu}_{\mathbf{f}_i}$ has no direct $\phi$ dependence. The squared-error term $-\frac{1}{2\sigma^2}\|\mathbf{y}_i-\boldsymbol{\mu}_{\mathbf{f}_i}\|^2$ and the inducing trace $\operatorname{Tr}(\mathbf{A}_{\delta,i}\mathbf{S}_{\delta_i}\mathbf{A}_{\delta,i}^\top)$ in $\mathcal{L}_i^{\mathrm{global}}$ therefore contribute zero direct $\phi$-derivatives in the standard scalar model unless the implementation parameterizes means or loadings differently. By the chain rule,
\begin{equation}
\label{eq:phi_gradient_explicit}
\begin{aligned}
&\left.
\frac{\partial \mathcal{L}_i^{\mathrm{global}}}{\partial \phi}
-\frac{\partial}{\partial \phi}\mathrm{KL}\bigl(q(\boldsymbol{\delta}_i)\,\|\,p(\boldsymbol{\delta}_i;\phi,\mathbf{K}_{gg})\bigr)
\right|_{\Theta_{\mathrm{global}}=\Theta_{\mathrm{global}}^{(r)}} \\
&=
\left.
\frac{\partial}{\partial \phi}\Bigl[
-\tfrac{1}{2\sigma^2}\operatorname{Tr}(\mathbf{K}_{\delta,ii} - \mathbf{Q}_{\delta,ii}) \right. \\
&\qquad\left.
-\mathrm{KL}\bigl(q(\boldsymbol{\delta}_i)\,\|\,p(\boldsymbol{\delta}_i;\phi,\mathbf{K}_{gg})\bigr)
\Bigr]\right|_{\Theta_{\mathrm{global}}=\Theta_{\mathrm{global}}^{(r)}},
\end{aligned}
\end{equation}
so that
\begin{equation}
\begin{aligned}
\mathcal{S}_{i,\eta_\phi}^{(r+1)}
&=
\left(
\left.
\frac{\partial \mathcal{L}_i^{\mathrm{global}}}{\partial \phi}
-\frac{\partial}{\partial \phi}
\mathrm{KL}\bigl(q(\boldsymbol{\delta}_i)\,\|\,p(\boldsymbol{\delta}_i;\phi,\mathbf{K}_{gg})\bigr)
\right|_{\Theta_{\mathrm{global}}=\Theta_{\mathrm{global}}^{(r)}}
\right)
\frac{\partial \phi}{\partial \eta_\phi}, \\
\frac{\partial \phi}{\partial \eta_\phi}&=\sigma(\eta_\phi)\in(0,1),
\end{aligned}
\end{equation}
with $\sigma(\cdot)$ the logistic function. Under the shared-basis parameterization $k_\delta=\phi k_g$, $\phi$ affects the global step primarily through the deviation residual trace $\operatorname{Tr}(\mathbf{K}_{\delta,ii}-\mathbf{Q}_{\delta,ii})$ and the deviation prior/KL term in~\eqref{eq:phi_gradient_explicit}. It can influence the fitted deviation mean only indirectly through subsequent local variational updates, because $q(\boldsymbol{\delta}_i)$ is reoptimized under the updated prior scale after $\phi$ is broadcast. In practice these gradients are computed by automatic differentiation within the federated block-coordinate loop. A more flexible extension would replace the shared $\phi$ with client-specific scales $\phi_i$ or a hierarchical prior over $\{\phi_i\}$; we leave this extension for future work to avoid increasing communication and server-side hyperparameter complexity.

\begin{algorithm}[t]
\caption{pFedHGP federated variational optimization (one communication round).}
\label{alg:pfedhgp}

\begin{algorithmic}[1]
\State \textbf{Server input:} $\Theta_{\mathrm{global}}^{(r)}$; \textbf{Clients $i=1,\ldots,T$:} private data $(\mathbf{X}_i,\mathbf{y}_i)$, local blocks $\Theta_{i,\mathrm{local}}^{(r)}$
\State \textbf{Broadcast} $\Theta_{\mathrm{global}}^{(r)}$ from server to all clients
\For{each client $i$ in parallel}
    \State Maximize local ELBO~\eqref{eq:local_elbo_new} over $\Theta_{i,\mathrm{local}}$ with $\Theta_{\mathrm{global}}^{(r)}$ \textbf{held fixed}; obtain $\Theta_{i,\mathrm{local}}^{(r+1)}$
    \State Compute $\widetilde{\mathcal{S}}_i^{(r+1)}$ from~\eqref{eq:Si_def} using $(\mathbf{X}_i,\mathbf{y}_i)$, $\Theta_{i,\mathrm{local}}^{(r+1)}$, and $\Theta_{\mathrm{global}}^{(r)}$
    \State \textbf{Transmit} $\widetilde{\mathcal{S}}_i^{(r+1)}$ to server (no raw data or local variational parameters)
\EndFor
\State \textbf{Server aggregation:} $\widetilde{\mathcal{S}}^{(r+1)} \gets \sum_{i=1}^T \widetilde{\mathcal{S}}_i^{(r+1)}$; form $\nabla_{\Theta_{\mathrm{global}}}\mathcal{L}$ via~\eqref{eq:global_gradient}
\State \textbf{Global update:} $\Theta_{\mathrm{global}}^{(r+1)} \gets \Theta_{\mathrm{global}}^{(r)} + \eta_{\mathrm{global}} \nabla_{\Theta_{\mathrm{global}}}\mathcal{L}$ (cf.~\eqref{eq:global_update})
\State \textbf{Broadcast} $\Theta_{\mathrm{global}}^{(r+1)}$ to clients for round $r+1$
\end{algorithmic}%

\end{algorithm}

Each round therefore alternates a \emph{local} block update with a \emph{global} aggregation step; within the local step the global parameters act as fixed conditioning values for the client sub-problem (Proposition~\ref{prop:local_elbo}), while the transmitted gradient $\widetilde{\mathcal{S}}_i^{(r+1)}$ couples updated local states back to the global block for the subsequent server update.

\paragraph{Summary and intuition (federated optimization).}
Algorithm~\ref{alg:pfedhgp} implements block coordinate descent: clients first maximize their local ELBO blocks with $\Theta_{\mathrm{global}}$ fixed, then transmit low-dimensional gradient summaries $\widetilde{\mathcal{S}}_i$ that already include each client's deviation-KL contribution, so the server updates the shared backbone---including $\eta_\phi$ (hence $\phi=\operatorname{softplus}(\eta_\phi)$) and $\mathbf{Z}_g$---using only aggregated gradients and the global inducing variational parameters, without pooling raw observations or local deviation posteriors.

\subsection{Implementation Details}
\label{sec:implementation}

\paragraph{Covariance factorization.}
We use Cholesky decomposition to factorize covariance matrices for numerical stability and to avoid explicit matrix inversion during variational updates. Specifically, for any covariance matrix $\mathbf{S}$, we compute the lower triangular Cholesky factor $\mathbf{L}$ such that $\mathbf{S} = \mathbf{L}\mathbf{L}^\top$. All covariance matrices are regularized with a fixed jitter $\epsilon = 10^{-6}$ before decomposition, i.e., we compute the Cholesky factorization of $\mathbf{K} + \epsilon\mathbf{I}$ where $\mathbf{K}$ is the original covariance matrix. This ensures numerical stability while maintaining the positive definiteness of the matrices.

\paragraph{Whitened variational parameterization.}
For numerical stability, we utilize the whitened variational strategy parameterization \citep{matthews2017scalable}. Given a prior distribution $p(\mathbf{u}) = \mathcal{N}(\mathbf{0}, \mathbf{K})$ with Cholesky factorization $\mathbf{K} = \mathbf{L}\mathbf{L}^\top$, we parameterize the inducing variables as $\mathbf{u} = \mathbf{L}\mathbf{v}$, where $\mathbf{v}$ are the whitened variational parameters. The variational distribution is then expressed as $q(\mathbf{v}) = \mathcal{N}(\tilde{\mathbf{m}}, \tilde{\mathbf{S}})$ with whitened mean $\tilde{\mathbf{m}}$ and covariance $\tilde{\mathbf{S}}$. The original variational parameters are recovered via $\mathbf{m} = \mathbf{L}\tilde{\mathbf{m}}$ and $\mathbf{S} = \mathbf{L}\tilde{\mathbf{S}}\mathbf{L}^\top$. This parameterization improves optimization by removing correlations in the variational posterior, leading to better-conditioned optimization problems and more stable gradient computations.

\paragraph{Hyperparameter optimization.}
All positive-constrained hyperparameters (lengthscales, variances) are optimized in an unconstrained space via a Softplus transformation. Specifically, for a positive hyperparameter $\theta > 0$, we introduce an unconstrained parameter $\tilde{\theta} \in \mathbb{R}$ and define $\theta = \log(1 + \exp(\tilde{\theta})) = \text{softplus}(\tilde{\theta})$. The gradient with respect to $\tilde{\theta}$ is computed using the chain rule: $\frac{\partial \mathcal{L}}{\partial \tilde{\theta}} = \frac{\partial \mathcal{L}}{\partial \theta} \cdot \sigma(\tilde{\theta})$, where $\sigma(\cdot)$ is the sigmoid function. This transformation ensures that the hyperparameters remain positive throughout optimization while allowing unconstrained gradient-based optimization algorithms (e.g., Adam, SGD) to be applied directly.

\paragraph{Inducing locations and kernel hyperparameters.}

We jointly optimize inducing point locations and kernel parameters via gradient ascent on the ELBO (marginal likelihood). The global component and the client-specific deviations share the same inducing locations $\mathbf{Z}_g$ to preserve a common basis, while each local component uses its own $\mathbf{Z}_i$ to capture client-specific residual structure. For the global kernel $k_g$ with hyperparameters $(\ell_g, \sigma_g^2)$ and the local kernel $k_i$ with hyperparameters $(\ell_i, \sigma_i^2)$, we optimize $\{\mathbf{Z}_g, \ell_g, \sigma_g^2, \eta_\phi, \{\mathbf{Z}_i, \ell_i, \sigma_i^2\}_{i=1}^T\}$ jointly with the variational parameters by maximizing the ELBO, with $\phi=\operatorname{softplus}(\eta_\phi)$ enforced by the transformation above. The gradients $\nabla_{\mathbf{Z}_g}\mathcal{L}$, $\nabla_{\ell_g}\mathcal{L}$, $\nabla_{\sigma_g^2}\mathcal{L}$, $\nabla_{\mathbf{Z}_i}\mathcal{L}$, $\nabla_{\ell_i}\mathcal{L}$, and $\nabla_{\sigma_i^2}\mathcal{L}$ are computed via automatic differentiation, enabling automatic adaptation of both global and local kernels.

\paragraph{VFE-style diagonal computations.}
For efficiency we employ diagonal computations for the trace terms in the VFE objective. Specifically, the trace regularization terms $\operatorname{Tr}(\mathbf{K}_{g,ii} - \mathbf{Q}_{g,ii})$, $\operatorname{Tr}(\mathbf{K}_{\delta,ii} - \mathbf{Q}_{\delta,ii})$, and $\operatorname{Tr}(\mathbf{K}_{i,ii} - \mathbf{Q}_{i,ii})$ are computed by summing only the diagonal elements: $\sum_{j=1}^{n_i} [\mathbf{K}_{\cdot,ii} - \mathbf{Q}_{\cdot,ii}]_{jj}$, where $[\cdot]_{jj}$ denotes the $j$-th diagonal element. Our inference is framed in the VFE objective: we retain the exact GP prior and rely on the trace of the residual covariance (which depends only on diagonal elements) rather than assuming a diagonal observation covariance structure. This approach provides computational efficiency while maintaining the exact prior and avoiding restrictive assumptions on the covariance structure.

\subsection{Practical Usage of pFedHGP}
\label{sec:usage}

Beyond regression and calibration tasks, pFedHGP naturally extends to classification and clustering applications by leveraging its hierarchical predictive distributions. Section~\ref{sec:classification} presents a maximum-evidence classification approach that uses the full Gaussian predictive log-likelihood under~\eqref{eq:pred_cov_classification} to assign test samples to client models, enabling federated fault diagnosis and anomaly detection while preserving privacy. Section~\ref{sec:clustering} introduces a structural comparison framework that computes client-specific projection operators in the inducing feature space, enabling unsupervised discovery of client groups with similar patterns without sharing raw observations.

\subsubsection{Classification via pFedHGP}
\label{sec:classification}

For each client \(i\), the variational posterior induces a Gaussian predictive law for observations at arbitrary inputs. At the training inputs \(\mathbf{X}_i\), we have already expressed the expected log likelihood in closed form in \eqref{eq:closed_form_ll} in terms of the hierarchical predictive mean \(\boldsymbol{\mu}_{\mathbf{f}_i}\), posterior covariance in each inducing block (\(\mathbf{S}_g\), \(\mathbf{S}_{\delta_i}\), \(\mathbf{S}_{u_i}\)), and the residual covariance \(\mathbf{R}_i\). The mean decomposes into global, client-specific deviation, and local contributions as in \eqref{eq:mean_decomposition}, and the residual trace \(\operatorname{Tr}(\mathbf{R}_i)\) decomposes additively across the same three components as in \eqref{eq:covariance_decomposition}.

For a batch of test inputs \(\mathbf{X}_\star\) and observations \(\mathbf{y}_\star\), we reuse exactly this structure. Let \(\mathbf{A}_{g,i}^{\star}\), \(\mathbf{A}_{\delta,i}^{\star}\), and \(\mathbf{A}_{i}^{\star}\) denote the interpolation matrices from Proposition~\ref{prop:low_rank} evaluated at \(\mathbf{X}_\star\) (so \(\mathbf{A}_{g,i}^{\star}=\mathbf{K}_g(\mathbf{X}_\star,\mathbf{Z}_g)\mathbf{K}_{gg}^{-1}\), \(\mathbf{A}_{\delta,i}^{\star}=\mathbf{A}_{g,i}^{\star}\) since \(k_\delta=\phi k_g\), and \(\mathbf{A}_{i}^{\star}=\mathbf{K}_i(\mathbf{X}_\star,\mathbf{Z}_i)\mathbf{K}_{ii}^{-1}\)). Under the trained variational posterior, client \(i\) assigns the Gaussian predictive law
\[
\mathbf{y}_\star \mid \mathbf{X}_\star, i
~\sim~
\mathcal{N}\!\bigl(\boldsymbol{\mu}_{\mathbf{f}_i}(\mathbf{X}_\star),\,\boldsymbol{\Sigma}_i(\mathbf{X}_\star)\bigr),
\]
with hierarchical predictive mean \(\boldsymbol{\mu}_{\mathbf{f}_i}(\mathbf{X}_\star)\) as in \eqref{eq:mean_decomposition} and predictive covariance
\begin{equation}
\label{eq:pred_cov_classification}
\boldsymbol{\Sigma}_i(\mathbf{X}_\star)
=
\mathbf{A}_{g,i}^{\star}\mathbf{S}_g(\mathbf{A}_{g,i}^{\star})^\top
+ \mathbf{A}_{\delta,i}^{\star}\mathbf{S}_{\delta_i}(\mathbf{A}_{\delta,i}^{\star})^\top
+ \mathbf{A}_{i}^{\star}\mathbf{S}_{u_i}(\mathbf{A}_{i}^{\star})^\top
+ \mathbf{R}_i(\mathbf{X}_\star) + \sigma^2\mathbf{I},
\end{equation}
where the first three terms are inducing-posterior covariance contributions and \(\mathbf{R}_i(\mathbf{X}_\star)\) is the VFE residual covariance at \(\mathbf{X}_\star\) not explained by the inducing variables (with trace decomposition as in \eqref{eq:covariance_decomposition}); the sum is the full predictive covariance and does not double-count uncertainty. The blocks in~\eqref{eq:pred_cov_classification} are the same uncertainty components that enter the training-time VFE objective~\eqref{eq:closed_form_ll}, but classification uses the \emph{marginal} Gaussian log-density at test inputs, not the trace-weighted expected log-likelihood from the ELBO. In particular, \eqref{eq:closed_form_ll} scales squared error and all covariance contributions by \(1/(2\sigma^2)\) and does not include the \(\boldsymbol{\Sigma}_i^{-1}\) quadratic form or \(\log|\boldsymbol{\Sigma}_i|\) term of a full multivariate predictive law.

Let \(n_\star := \dim(\mathbf{y}_\star)\). The classification score is the Gaussian predictive log-likelihood
\begin{equation}
\label{eq:cls-ll}
\begin{aligned}
\ell_i(\mathbf{X}_\star,\mathbf{y}_\star)
&:= \log \mathcal{N}\!\bigl(\mathbf{y}_\star;\,\boldsymbol{\mu}_{\mathbf{f}_i}(\mathbf{X}_\star),\,\boldsymbol{\Sigma}_i(\mathbf{X}_\star)\bigr) \\
&= -\frac{1}{2}\Bigl[
\bigl(\mathbf{y}_\star - \boldsymbol{\mu}_{\mathbf{f}_i}(\mathbf{X}_\star)\bigr)^\top
\boldsymbol{\Sigma}_i(\mathbf{X}_\star)^{-1}
\bigl(\mathbf{y}_\star - \boldsymbol{\mu}_{\mathbf{f}_i}(\mathbf{X}_\star)\bigr)
+ \log\bigl|\boldsymbol{\Sigma}_i(\mathbf{X}_\star)\bigr|
+ n_\star \log(2\pi)
\Bigr].
\end{aligned}
\end{equation}
We evaluate~\eqref{eq:cls-ll} in practice via Cholesky factorization of \(\boldsymbol{\Sigma}_i(\mathbf{X}_\star)\).

We then use a standard model-evidence decision rule, treating each client model as a class-conditional generator. The predicted label for \((\mathbf{X}_\star,\mathbf{y}_\star)\) is chosen by maximum predictive log likelihood:
\begin{equation}
\label{eq:cls-rule}
\hat{c} \;=\; \arg\max_{i\in\{1,\dots,T\}} ~ \ell_i(\mathbf{X}_\star,\mathbf{y}_\star).
\end{equation}
Because \(\boldsymbol{\mu}_{\mathbf{f}_i}(\mathbf{X}_\star)\), \(\boldsymbol{\Sigma}_i(\mathbf{X}_\star)\), and \(\mathbf{R}_i(\mathbf{X}_\star)\) inherit the hierarchical decompositions in \eqref{eq:mean_decomposition} and \eqref{eq:covariance_decomposition}, the score \(\ell_i\) automatically discounts patterns shared through the global component while emphasizing client-specific offsets and local structure, which is particularly valuable in heterogeneous classification settings.

\subsubsection{Clustering via pFedHGP}
\label{sec:clustering}
For structural comparison across clients, each client is represented by a projection operator in the inducing feature space that captures the subspaces effectively spanned by its offset and local components. Using only \(\mathbf{X}_i\) and the trained kernels (privacy preserved), define the offset operator \(A_{\delta,i}=K_g(\mathbf{Z}_g,\mathbf{X}_i)\,(K_g(\mathbf{X}_i,\mathbf{X}_i)+\sigma^2 I_{n_i})^{-1}\,K_g(\mathbf{X}_i,\mathbf{Z}_g)\in\mathbb{R}^{M\times M}\), which leverages the global kernel \(K_g\) since the client-specific offset \(\delta_i\) does not possess an independent kernel. The local operator is \(A_{\mathrm{loc},i}=K_i(\mathbf{Z}_i,\mathbf{X}_i)\,(K_i(\mathbf{X}_i,\mathbf{X}_i)+\sigma^2 I_{n_i})^{-1}\,K_i(\mathbf{X}_i,\mathbf{Z}_i)\in\mathbb{R}^{M_i\times M_i}\). The joint personalized operator is given by \(A_i=\mathrm{blockdiag}(A_{\delta,i},A_{\mathrm{loc},i})\), and its trace-normalized version is \(\widehat A_i=A_i/\mathrm{tr}(A_i)\). Dissimilarity between clients is quantified by the Frobenius distance \(\rho(i,j)=\|\widehat A_i-\widehat A_j\|_F^2\), and the dissimilarity matrix \(\big[\rho(i,j)\big]_{i,j=1}^T\) can be fed directly into standard clustering routines. Since \(\widehat A_i\) is computed locally from \(\mathbf{X}_i\) and kernel evaluations, clustering requires no exposure of raw observations and remains coherent with the federated training pipeline.

\subsection{ILMM Formulation of pFedHGP}
\label{sec:ilmm_formulation}

We connect pFedHGP with the \emph{Instantaneous Linear Mixing Model} (ILMM) \citep{bruinsma2020scalable}, a framework for constructing multi-output Gaussian processes by linearly mixing latent Gaussian processes. In function space, the hierarchical decomposition from Assumption~\ref{assump:generative_model} (where $f_i^{\mathrm{tot}}(x) = f_g(x) + f_{\delta,i}(x) + f_i(x)$) can be recast as an ILMM by stacking all clients' latent functions into a $T$-dimensional output vector $\mathbf{f}^{\mathrm{tot}}(x)$ and a $(2T+1)$-dimensional latent vector $\mathbf{x}(x) = [f_g(x), f_{\delta,1}(x), \dots, f_{\delta,T}(x), f_1(x), \dots, f_T(x)]^\top$. The mapping from latent to output processes is then linear:
\begin{equation}
\label{eq:ilmm_mixing}
\mathbf{f}^{\mathrm{tot}}(x) = H\,\mathbf{x}(x), \quad H = \bigl[\,\mathbf{1}_T \;\big|\; I_T \;\big|\; I_T\,\bigr] \in \mathbb{R}^{T\times(2T+1)},
\end{equation}
where $\mathbf{1}_T$ is the $T$-vector of ones and $I_T$ is the $T\times T$ identity matrix. With observation noise $\Sigma = \sigma^2 I_T$, this yields an ILMM with $p=T$ outputs and $m=2T+1$ latent processes. The choice $k_\delta=\phi k_g$ (from Assumption~\ref{assump:generative_model}) ensures that client-specific deviations $f_{\delta,i}$ live in the same latent subspace as the shared global process $f_g$.

In inducing-variable space, Proposition~\ref{prop:low_rank} reveals a similar linear mixing structure through the low-rank expansion~\eqref{eq:low_rank_expansion}, where the shared basis $\mathbf{A}_{g,i}$ (common to $\mathbf{u}_g$ and $\boldsymbol{\delta}_i$) and client-specific local bases $\mathbf{A}_{i}$ organize the hierarchical decomposition. Under our VFE implementation, the residual covariance $\mathbf{R}_i$ (Proposition~\ref{prop:covariance}) is incorporated through trace regularization terms in the ELBO (Theorem~\ref{prop:elbo_decomp}), rather than directly modifying the observation model. This ILMM perspective embeds pFedHGP within the multi-output GP mixing-model hierarchy while maintaining the privacy-preserving federated structure.

\paragraph{Scalar-to-multi-sensor vector extension.}
The same ILMM viewpoint extends pFedHGP from scalar client responses to synchronized multi-sensor observations. Suppose client \(i\) observes
\(\mathbf{y}_i(x)=(y_{i1}(x),\ldots,y_{iQ}(x))^\top\in\mathbb{R}^Q\). A vector-valued pFedHGP preserves the same hierarchical decomposition by representing the \(Q\)-dimensional global, deviation, and local components through latent GP factors. We denote ILMM \emph{loading} matrices by \(B_{\cdot}\) to distinguish them from sparse-GP \emph{interpolation} matrices \(\mathbf{A}_{\cdot,i}\) in Section~\ref{sec:inducing_variables}:
\begin{equation}
\label{eq:multi_sensor_ilmm_main}
\mathbf{y}_i(x)
=
B_g\mathbf{h}_g(x)
+
B_{\delta,i}\mathbf{h}_{\delta,i}(x)
+
B_{l,i}\mathbf{h}_{l,i}(x)
+
\boldsymbol{\epsilon}_i(x),
\qquad
\boldsymbol{\epsilon}_i(x)\sim\mathcal{N}(\mathbf{0},\Sigma_\epsilon),
\end{equation}
where \(B_g\in\mathbb{R}^{Q\times R_g}\), \(B_{\delta,i}\in\mathbb{R}^{Q\times R_\delta}\), and \(B_{l,i}\in\mathbb{R}^{Q\times R_l}\) map latent global, deviation, and local GP factors to the \(Q\) sensor channels. The latent vectors \(\mathbf{h}_g(x)\), \(\mathbf{h}_{\delta,i}(x)\), and \(\mathbf{h}_{l,i}(x)\) represent shared population-level factors, client-specific systematic calibration factors, and local residual factors, respectively.

This construction induces the cross-client and cross-sensor covariance
\begin{equation}
\label{eq:multi_sensor_cov_main}
\begin{aligned}
\operatorname{Cov}\{\mathbf{y}_i(x),\mathbf{y}_j(x')\}
&=
B_gK_g(x,x')B_g^\top \\
&\quad+
\mathbf{1}[i=j]B_{\delta,i} K_\delta(x,x')B_{\delta,i}^\top \\
&\quad+
\mathbf{1}[i=j]B_{l,i}K_l(x,x')B_{l,i}^\top
+
\mathbf{1}[i=j]\Sigma_\epsilon\mathbf{1}[x=x'] ,
\end{aligned}
\end{equation}
where \(K_g(x,x')=\operatorname{Cov}\{\mathbf{h}_g(x),\mathbf{h}_g(x')\}\), \(K_\delta(x,x')=\operatorname{Cov}\{\mathbf{h}_{\delta,i}(x),\mathbf{h}_{\delta,i}(x')\}\), and \(K_l(x,x')=\operatorname{Cov}\{\mathbf{h}_{l,i}(x),\mathbf{h}_{l,i}(x')\}\). Equation~\eqref{eq:multi_sensor_cov_main} shows that cross-sensor dependence is generated by latent trajectories projected through the loading matrices. The global term is shared across clients, whereas the deviation, local, and noise terms are client-specific. 

\paragraph{Summary and intuition (ILMM).}
The ILMM viewpoint supplies a positive-semidefinite \(Q\times Q\) covariance construction for correlated sensor channels; it does not introduce a separate scalar computational shortcut. Scalability follows from sparse variational inducing variables, trace regularization, and federated aggregation of global statistics.

\section{Numerical Study}
\label{sec:experiments}

This section evaluates pFedHGP in a controlled multi-output federated regression problem. The experiment is designed to test four properties that are central to the proposed framework: predictive accuracy, recovery of cross-output dependence, uncertainty calibration, and robustness to violations of the assumed additive hierarchy. Each client observes multiple correlated output channels over a common input domain. The ground-truth data-generating process is known, which permits direct evaluation of both prediction quality and the learned output covariance structure. All reported multi-output results are based on \textbf{30 independent runs}. In each run we regenerate the latent component trajectories, draw new component loading matrices, resample client training inputs, reinitialize every fitted model, and draw fresh observation noise. Metrics are averaged over the \(T=6\) clients within each run; tables report run-level means with \(95\%\) confidence intervals computed across the 30 runs.

\subsection{Multi-output Synthetic Benchmark}
\label{sec:data_generation}

We consider \(T=6\) clients and \(Q=4\) output channels over the input domain \([0,10]\). For each independent run, client \(i\in\{1,\ldots,T\}\) receives \(n_{\mathrm{train}}=50\) training inputs sampled uniformly from \([0,10]\), and all clients are evaluated on a common dense test grid with \(n_{\mathrm{test}}=150\) inputs. The observation at input \(x\) is a vector
\(\mathbf{y}_i(x)=(y_{i1}(x),\ldots,y_{iQ}(x))^\top\in\mathbb{R}^Q\), generated by the additive multi-output hierarchy
\begin{equation}
\label{eq:multioutput_generation}
\mathbf{y}_i(x)
=
\underbrace{
\mathbf{f}_g(x)
+\mathbf{f}_{\delta,i}(x)
+\mathbf{f}_i(x)
}_{\mathbf{y}^{\mathrm{clean}}_i(x)}
+\boldsymbol{\epsilon}_i(x),
\qquad
\boldsymbol{\epsilon}_i(x)\sim N(\mathbf{0},\sigma^2 I_Q),
\end{equation}
where \(\sigma=0.05\). The components \(\mathbf{f}_g\), \(\mathbf{f}_{\delta,i}\), and \(\mathbf{f}_i\) represent, respectively, the shared global function, the systematic client-specific deviation, and the local site-specific variation. Cross-output dependence is introduced within these vector-valued component processes. All component processes are generated from RBF covariance functions of the form
\begin{equation}
\label{eq:global_rbf_kernel}
k(x,x')
=
\tau^2
\exp\left\{-\frac{(x-x')^2}{2\ell^2}\right\}.
\end{equation}
To induce correlated four-dimensional outputs, each vector-valued component is formed by mixing \(R=2\) independent scalar GP draws into the \(Q=4\) output channels. Concretely, for component \(c\in\{g,\delta,l\}\) we draw \(\mathbf{h}_c(x)=(h_{c,1}(x),h_{c,2}(x))^\top\) and set
\[
\mathbf{f}_c(x)=B_c\,\mathbf{h}_c(x),
\qquad
B_c\in\mathbb{R}^{Q\times R},
\]
with independent standard-normal entries in each \(B_c\) resampled at every run (and client-specific \(B_{\delta,i}\), \(B_{l,i}\) drawn independently in the deviation and local components). The two global latent draws use lengthscales \(\ell_{g,1}=1.8\) and \(\ell_{g,2}=2.5\), with output scales \(\tau_{g,1}=1.0\) and \(\tau_{g,2}=0.7\). The deviation latents use the same smooth RBF lengthscales as the global component, with simulator deviation scale \(\phi_{\mathrm{sim}}=1.20\). The local latents use shorter lengthscales \(\ell_{l,1}=0.25\) and \(\ell_{l,2}=0.45\), with output scales \(\tau_{l,1}=1.0\) and \(\tau_{l,2}=0.8\).

\subsection{Fitted ILMM--pFedHGP configuration and training}
\label{sec:exp_implementation}

All fitted methods target the ILMM observation model in~\eqref{eq:multi_sensor_ilmm_main} with \(Q=4\) and latent ranks \(R_g=R_\delta=R_l=2\), matching the two-latent-draw simulator above. For each scalar latent dimension we use the sparse hierarchical pFedHGP parameterization from Sections~\ref{sec:inducing_variables}--\ref{sec:algorithm}: shared global inducing locations \(\mathbf{Z}_g\in\mathbb{R}^{M\times 1}\) with \(M=25\) points initialized on a uniform grid over \([0,10]\), client-specific local inducing grids \(\mathbf{Z}_i\in\mathbb{R}^{M_i\times 1}\) with \(M_i=25\), squared-exponential kernels \(k_g\) and \(\{k_i\}\), and deviation layer \(k_\delta=\phi k_g\) with population scale \(\phi\) learned on the server. The loading matrices are \emph{not} fixed to the simulator draws: we treat \(B_g\in\mathbb{R}^{4\times 2}\) as a globally shared learnable block and \(\{B_{\delta,i},B_{l,i}\}\) as client-specific learnable blocks, all optimized jointly with the variational parameters and kernel hyperparameters by gradient ascent on the federated ELBO. Observation noise variance is learned through a softplus reparameterization with initialization \(\sigma^2=\sigma_{\mathrm{sim}}^2=0.05^2\). Initial kernel lengthscales are set to \(\ell_g=2.0\), \(\ell_i=0.35\), and output variances to \(1.0\); inducing locations remain learnable as in Section~\ref{sec:implementation}.

Training follows Algorithm~\ref{alg:pfedhgp} for \(R_{\mathrm{comm}}=20\) synchronous communication rounds. In round \(r\), each client first runs \(S_{\mathrm{loc}}=80\) Adam steps (learning rate \(\eta_{\mathrm{loc}}=0.1\)) on its local ELBO block with \(\Theta_{\mathrm{global}}^{(r)}\) held fixed, then transmits \(\widetilde{\mathcal{S}}_i^{(r+1)}\) from~\eqref{eq:Si_def}. The server forms \(\nabla_{\Theta_{\mathrm{global}}}\mathcal{L}\) via~\eqref{eq:global_gradient} (the \(\eta_\phi\) block is the corresponding component of \(\sum_i \widetilde{\mathcal{S}}_i^{(r+1)}\), cf.~\eqref{eq:eta_phi_update}) and applies one Adam step with learning rate \(\eta_{\mathrm{global}}=0.1\). Reduced models use the same protocol: \textbf{No-deviation} removes \(\mathbf{f}_{\delta,i}\) and \(B_{\delta,i}\); \textbf{No-local} removes \(\mathbf{f}_i\) and \(B_{l,i}\); \textbf{Global-only} and \textbf{Local-only} retain only the corresponding hierarchical blocks; \textbf{pFedGP} keeps client-specific multi-output blocks with FedAvg-style synchronization of shared input-kernel hyperparameters but without the explicit global/deviation/local ILMM decomposition.

At test time, client \(i\) returns the joint predictive Gaussian
\(\mathcal{N}(\widehat{\boldsymbol{\mu}}_i(x),\widehat{\Sigma}_i(x))\) obtained by composing the hierarchical predictive mean and covariance in Section~\ref{sec:classification} with the learned ILMM loadings. Let \(\mathbf{A}_{g}^{\star}(x)\in\mathbb{R}^{R_g\times M}\), \(\mathbf{A}_{\delta,i}^{\star}(x)\in\mathbb{R}^{R_\delta\times M}\), and \(\mathbf{A}_{l,i}^{\star}(x)\in\mathbb{R}^{R_l\times M_i}\) denote the rows of the interpolation matrices in Proposition~\ref{prop:low_rank} evaluated at the scalar input \(x\) (one row per latent GP coordinate, as in~\eqref{eq:pred_cov_classification} at a single test location). Then
\begin{equation}
\label{eq:multioutput_pred_mean}
\widehat{\boldsymbol{\mu}}_i(x)
=
B_g\,\widehat{\mathbf{h}}_g(x)
+
B_{\delta,i}\,\widehat{\mathbf{h}}_{\delta,i}(x)
+
B_{l,i}\,\widehat{\mathbf{h}}_{l,i}(x),
\end{equation}
\begin{equation}
\label{eq:multioutput_pred_cov}
\widehat{\Sigma}_i(x)
=
B_g\,\widehat{\mathbf{S}}_g(x)\,B_g^\top
+
B_{\delta,i}\,\widehat{\mathbf{S}}_{\delta,i}(x)\,B_{\delta,i}^\top
+
B_{l,i}\,\widehat{\mathbf{S}}_{l,i}(x)\,B_{l,i}^\top
+
\widehat{\mathbf{R}}_i(x)+\sigma^2 I_Q,
\end{equation}
with inducing-posterior latent summaries
\(\widehat{\mathbf{h}}_g(x)=\mathbf{A}_{g}^{\star}(x)\mathbf{m}_g\),
\(\widehat{\mathbf{h}}_{\delta,i}(x)=\mathbf{A}_{\delta,i}^{\star}(x)\mathbf{m}_{\delta_i}\),
\(\widehat{\mathbf{h}}_{l,i}(x)=\mathbf{A}_{l,i}^{\star}(x)\mathbf{m}_{u_i}\),
and
\(\widehat{\mathbf{S}}_g(x)=\mathbf{A}_{g}^{\star}(x)\mathbf{S}_g\{\mathbf{A}_{g}^{\star}(x)\}^{\top}\),
\(\widehat{\mathbf{S}}_{\delta,i}(x)=\mathbf{A}_{\delta,i}^{\star}(x)\mathbf{S}_{\delta_i}\{\mathbf{A}_{\delta,i}^{\star}(x)\}^{\top}\),
\(\widehat{\mathbf{S}}_{l,i}(x)=\mathbf{A}_{l,i}^{\star}(x)\mathbf{S}_{u_i}\{\mathbf{A}_{l,i}^{\star}(x)\}^{\top}\).
Here \(\widehat{\mathbf{S}}_{\cdot}(x)\) is the \emph{inducing-posterior} covariance contribution in latent space (analogous to the \(\mathbf{A}_{\cdot,i}^{\star}\mathbf{S}_{\cdot}(\mathbf{A}_{\cdot,i}^{\star})^{\top}\) blocks in~\eqref{eq:pred_cov_classification}), not the full latent predictive covariance.
The term \(\widehat{\mathbf{R}}_i(x)\in\mathbb{R}^{Q\times Q}\) is the VFE residual covariance at \(x\)---uncertainty not explained by the inducing variables, with the same additive global/deviation/local decomposition as in~\eqref{eq:covariance_decomposition}---mapped to the output channels through \(\{B_g,B_{\delta,i},B_{l,i}\}\).
Adding \(\widehat{\mathbf{R}}_i(x)\) therefore does not double-count the inducing-posterior terms carried by \(\widehat{\mathbf{S}}_{\cdot}(x)\).

\subsection{Assumption-mismatch Scenarios}
\label{sec:scenarios_methods}

We evaluate one matched scenario and three deliberately misspecified scenarios. These scenarios isolate different ways in which the fitted pFedHGP assumptions may fail in practice.

\paragraph{Scenario A: matched hierarchy.}
The data follow the additive structure in \eqref{eq:multioutput_generation}. The global, deviation, and local component processes are mutually independent; the deviation process is smooth; and the local process has short lengthscale. This scenario tests whether the fitted hierarchy can recover the intended multi-output decomposition when the modeling assumptions are approximately correct.

\paragraph{Scenario B: deviation-kernel mismatch.}
The fitted model retains an RBF deviation component, but the true deviation process contains an additional periodic structure. Specifically, the deviation kernel is
\begin{equation}
\label{eq:periodic_rbf_kernel}
k_{\delta}^{\mathrm{true}}(x,x')
=
0.65\,\tau_{\delta}^2
\exp\left\{-\frac{(x-x')^2}{2\ell_{\delta}^2}\right\}
+0.35\,\tau_{\delta}^2
\exp\left\{
-\frac{2\sin^2\{\pi |x-x'|/3\}}{0.8^2}
\right\}.
\end{equation}
This tests robustness to an incorrect functional form for systematic client drift.

\paragraph{Scenario C: non-additive interaction.}
The true clean signal includes a multiplicative interaction between the shared global component and the client-deviation component:
\begin{equation}
\label{eq:nonadditive_generation}
\mathbf{y}^{\mathrm{clean}}_i(x)
=
\mathbf{f}_g(x)
+\mathbf{f}_{\delta,i}(x)
+\mathbf{f}_i(x)
+\gamma_i\{\mathbf{f}_g(x)\odot \mathbf{f}_{\delta,i}(x)\},
\end{equation}
where \(\odot\) denotes elementwise multiplication and \(\gamma_i\sim\operatorname{Unif}(0.35,0.60)\). This setting violates the additive pFedHGP decomposition.

\paragraph{Scenario D: correlated hierarchical components.}
The fitted model assumes independent hierarchical components, whereas the true local residual is partially correlated with the deviation process:
\begin{equation}
\label{eq:correlated_component_generation}
\mathbf{f}_i(x)
=
\rho \mathbf{f}_{\delta,i}(x)
+\sqrt{1-\rho^2}\,\widetilde{\mathbf{f}}_i(x),
\qquad
\rho=0.5,
\end{equation}
where \(\widetilde{\mathbf{f}}_i\) is an independent short-lengthscale local GP. This scenario evaluates sensitivity to dependence between the deviation and local residual levels.

\subsection{Compared Models}
\label{sec:compared_methods}

We compare Full pFedHGP with five alternatives that represent different modeling assumptions and degrees of hierarchical structure.

\paragraph{No-deviation.}
This ablated variant removes the client-specific deviation component \(\mathbf{f}_{\delta,i}\) and fits \(\mathbf{f}_g+\mathbf{f}_i\). It tests whether systematic client-specific calibration provides additional value beyond a simpler global-plus-local decomposition.

\paragraph{No-local.}
This ablated variant removes the local residual component \(\mathbf{f}_i\) and fits \(\mathbf{f}_g+\mathbf{f}_{\delta,i}\). It evaluates the effect of excluding short-scale client-specific residual variation from the hierarchy.

\paragraph{Global-only.}
This model fits only the shared global component \(\mathbf{f}_g\), pooling all clients without personalization. It represents the assumption that all clients follow a common population-level process.

\paragraph{Local-only.}
This model fits client-specific private multi-output GPs without shared global or systematic deviation structure. It represents the opposite assumption that each client should be modeled independently, without borrowing information across clients.

\paragraph{pFedGP.}
This baseline uses the personalized federated Gaussian process method proposed by \citet{achituve2021personalized}. In our multi-output benchmark, each client maintains a personal multi-output GP block, and model parameters are synchronized by FedAvg-style averaging. Unlike pFedHGP, this baseline does not decompose the response into explicit global, deviation, and local components.

\subsection{Evaluation Protocol and Metrics}
\label{sec:evaluation_protocol}

Each scenario is run for \(30\) independent random seeds. Each client has \(50\) training observations and is evaluated on a common grid of \(150\) test inputs. Point prediction is measured by the client-averaged root mean squared error (RMSE):
\begin{equation}
\label{eq:rmse_metric}
\operatorname{RMSE}
=
\frac{1}{T}\sum_{i=1}^{T}
\left\{
\frac{1}{n_{\mathrm{test}}Q}
\sum_{j=1}^{n_{\mathrm{test}}}
\|\mathbf{y}_i(x_j)-\widehat{\boldsymbol{\mu}}_i(x_j)\|_2^2
\right\}^{1/2}.
\end{equation}

Probabilistic fit is measured by the client-averaged multivariate negative log-likelihood (NLL). If
\(\widehat{\boldsymbol{\mu}}_{ij}\) and \(\widehat{\Sigma}_{ij}\) are the predictive mean and full \(Q\times Q\) covariance block for client \(i\) at test input \(x_j\), then
\begin{equation}
\label{eq:multioutput_nll}
\operatorname{NLL}
=
\frac{1}{T}\sum_{i=1}^{T}
\frac{1}{n_{\mathrm{test}}}
\sum_{j=1}^{n_{\mathrm{test}}}
\frac{1}{2}
\left[
Q\log(2\pi)
+\log |\widehat{\Sigma}_{ij}|
+\{\mathbf{y}_i(x_j)-\widehat{\boldsymbol{\mu}}_{ij}\}^{\top}
\widehat{\Sigma}_{ij}^{-1}
\{\mathbf{y}_i(x_j)-\widehat{\boldsymbol{\mu}}_{ij}\}
\right].
\end{equation}

We also report the Gaussian closed-form continuous ranked probability score (CRPS). For output \(q\), let
\(\widehat{\mu}_{ijq}\) and \(\widehat{\sigma}_{ijq}^2\) denote the marginal predictive mean and variance, and define
\(z_{ijq}=(y_{iq}(x_j)-\widehat{\mu}_{ijq})/\widehat{\sigma}_{ijq}\). Then
\begin{equation}
\label{eq:crps_metric}
\operatorname{CRPS}
=
\frac{1}{T n_{\mathrm{test}}Q}
\sum_{i=1}^{T}\sum_{j=1}^{n_{\mathrm{test}}}\sum_{q=1}^{Q}
\widehat{\sigma}_{ijq}
\left[
z_{ijq}\{2\Phi(z_{ijq})-1\}
+2\varphi(z_{ijq})
-\frac{1}{\sqrt{\pi}}
\right],
\end{equation}
where \(\Phi(\cdot)\) and \(\varphi(\cdot)\) denote the standard normal cumulative distribution function and probability density function, respectively (distinct from the deviation scale \(\phi\)).

We further report empirical \(95\%\) interval coverage and \(95\%\) interval width. Let \(z_{0.95}=\Phi^{-1}(0.975)\) denote the standard normal critical value for a two-sided \(95\%\) interval:
\begin{align}
\operatorname{Cov}_{0.95}
&=
\frac{1}{T n_{\mathrm{test}}Q}
\sum_{i=1}^{T}
\sum_{j=1}^{n_{\mathrm{test}}}
\sum_{q=1}^{Q}
\mathbb{I}\left\{
|y_{iq}(x_j)-\widehat{\mu}_{ijq}|
\le
z_{0.95}\widehat{\sigma}_{ijq}
\right\},\\
\operatorname{Width}_{0.95}
&=
\frac{1}{T n_{\mathrm{test}}Q}
\sum_{i=1}^{T}
\sum_{j=1}^{n_{\mathrm{test}}}
\sum_{q=1}^{Q}
2z_{0.95}\widehat{\sigma}_{ijq}.
\end{align}

To evaluate recovery of cross-output dependence, we first form the ground-truth output covariance on the common test grid \(\mathcal{X}_{\mathrm{test}}\):
\begin{equation}
\label{eq:true_output_covariance}
\Sigma^{\mathrm{true}}_{y,i}
=
\operatorname{Cov}_{x\in\mathcal{X}_{\mathrm{test}}}
\bigl\{\mathbf{y}^{\mathrm{clean}}_i(x)\bigr\}
+\sigma_{\mathrm{sim}}^2 I_Q .
\end{equation}
For each fitted method, predictive distributions are joint Gaussians over the \(Q=4\) channels at every test input. Let \(\widehat{\boldsymbol{\mu}}_{ij}\in\mathbb{R}^Q\) and \(\widehat{\Sigma}_{ij}\in\mathbb{R}^{Q\times Q}\) denote the predictive mean and full covariance block for client \(i\) at test input \(x_j\in\mathcal{X}_{\mathrm{test}}\). We aggregate these blocks into a model-implied output covariance
\begin{equation}
\label{eq:model_implied_output_covariance}
\widehat{\Sigma}_{y,i}
=
\operatorname{Cov}_{x\in\mathcal{X}_{\mathrm{test}}}
\bigl\{\widehat{\boldsymbol{\mu}}_i(x)\bigr\}
+
\frac{1}{n_{\mathrm{test}}}
\sum_{j=1}^{n_{\mathrm{test}}}
\widehat{\Sigma}_{ij},
\end{equation}
and report the relative Frobenius error
\begin{equation}
\label{eq:cov_error}
\operatorname{CovErr}
=
\frac{1}{T}\sum_{i=1}^{T}
\frac{
\|\widehat{\Sigma}_{y,i}-\Sigma^{\mathrm{true}}_{y,i}\|_F
}{
\|\Sigma^{\mathrm{true}}_{y,i}\|_F
}.
\end{equation}
Lower \(\operatorname{CovErr}\) indicates more accurate recovery of cross-output covariance structure. Lower NLL, CRPS, RMSE, and width are also better, while coverage should be close to the nominal level.

\subsection{Results}
\label{sec:experiment_results}

\begin{figure}[ht]
\centering
\includegraphics[width=0.9\textwidth]{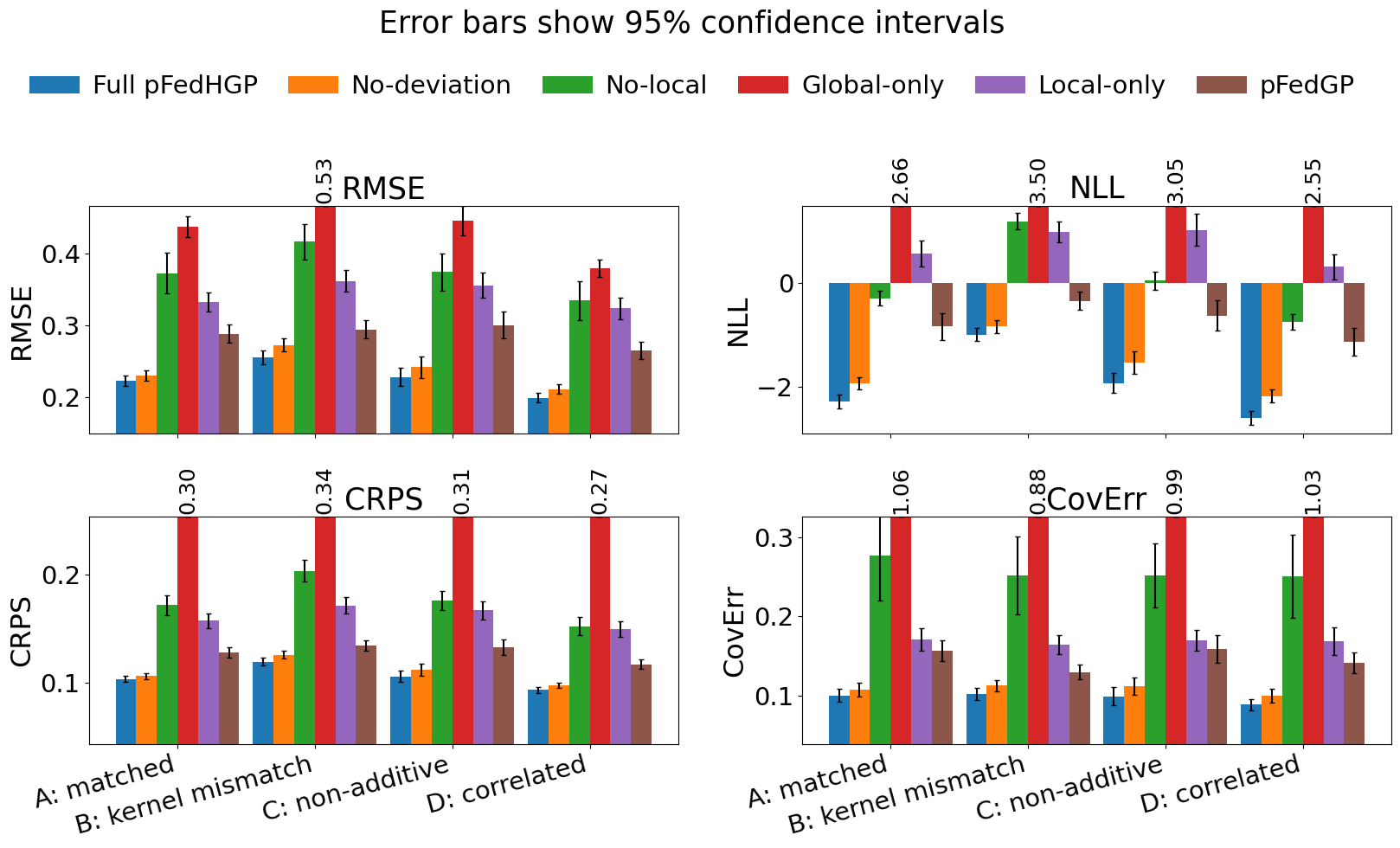}
\caption{Robustness summary across matched, deviation-kernel mismatch, non-additive, and correlated-component scenarios. Error bars denote 95\% confidence intervals computed from the results of 30 independent runs. Bars exceeding the plotting range are clipped for readability, with annotations showing their true values.
}
\label{fig:robustness_summary}
\end{figure}

Table~\ref{tab:robustness} reports the quantitative comparison, and Figure~\ref{fig:robustness_summary} summarizes the main accuracy, probabilistic, and covariance-recovery metrics graphically. Lower values are better for RMSE, NLL, CRPS, CovErr, and 95\% Width, while 95\% coverage should be close to the nominal level. We organize the discussion around three empirical patterns: the value of the full hierarchy, robustness and covariance recovery under misspecification, and the coverage--sharpness behavior of the predictive uncertainty.

\begin{table}[ht]
\centering
\caption{Multi-output performance across matched and assumption-mismatch scenarios. Entries report run-level means with 95\% confidence intervals in brackets.}
\label{tab:robustness}
\resizebox{\textwidth}{!}{%
\begin{tabular}{llcccccc}\toprule
Scenario & Model & RMSE & NLL & CRPS & CovErr & 95\% Width & 95\% Cov. \\ \midrule
A: matched & Full pFedHGP & \textbf{0.223 [0.216, 0.230]} & \textbf{-2.275 [-2.405, -2.145]} & \textbf{0.104 [0.101, 0.107]} & \textbf{0.100 [0.092, 0.108]} & \textbf{0.765 [0.702, 0.829]} & 0.938 [0.929, 0.948] \\
 & No-deviation & 0.231 [0.223, 0.238] & -1.922 [-2.036, -1.807] & 0.107 [0.104, 0.109] & 0.107 [0.098, 0.117] & \textbf{0.765 [0.709, 0.820]} & 0.938 [0.930, 0.947] \\
 & No-local & 0.373 [0.345, 0.401] & -0.301 [-0.441, -0.160] & 0.172 [0.162, 0.181] & 0.277 [0.220, 0.333] & 1.254 [1.139, 1.369] & 0.943 [0.935, 0.951] \\
 & Global-only & 0.438 [0.423, 0.453] & 2.655 [2.529, 2.781] & 0.297 [0.288, 0.307] & 1.063 [0.892, 1.234] & 1.845 [1.723, 1.968] & 0.943 [0.936, 0.950] \\
 & Local-only & 0.333 [0.320, 0.346] & 0.553 [0.309, 0.797] & 0.157 [0.151, 0.164] & 0.171 [0.156, 0.185] & 1.264 [1.137, 1.391] & 0.941 [0.932, 0.950] \\
 & pFedGP & 0.289 [0.276, 0.301] & -0.839 [-1.094, -0.584] & 0.128 [0.124, 0.133] & 0.157 [0.144, 0.170] & 0.931 [0.859, 1.003] & 0.942 [0.934, 0.950] \\
\addlinespace
B: kernel mismatch & Full pFedHGP & \textbf{0.256 [0.247, 0.265]} & \textbf{-0.996 [-1.119, -0.873]} & \textbf{0.120 [0.116, 0.123]} & \textbf{0.102 [0.094, 0.110]} & \textbf{0.919 [0.845, 0.993]} & 0.944 [0.935, 0.953] \\
 & No-deviation & 0.273 [0.264, 0.282] & -0.837 [-0.962, -0.712] & 0.126 [0.122, 0.130] & 0.112 [0.105, 0.120] & 0.939 [0.864, 1.013] & 0.940 [0.930, 0.950] \\
 & No-local & 0.417 [0.392, 0.442] & 1.169 [1.011, 1.326] & 0.203 [0.194, 0.213] & 0.252 [0.202, 0.301] & 1.506 [1.384, 1.627] & 0.941 [0.932, 0.949] \\
 & Global-only & 0.535 [0.521, 0.549] & 3.504 [3.420, 3.589] & 0.339 [0.332, 0.347] & 0.881 [0.781, 0.980] & 2.187 [2.076, 2.298] & 0.942 [0.936, 0.948] \\
 & Local-only & 0.362 [0.347, 0.378] & 0.974 [0.774, 1.173] & 0.172 [0.164, 0.179] & 0.164 [0.153, 0.176] & 1.416 [1.272, 1.560] & 0.941 [0.930, 0.951] \\
 & pFedGP & 0.295 [0.282, 0.308] & -0.354 [-0.526, -0.181] & 0.135 [0.130, 0.139] & 0.130 [0.121, 0.139] & 0.986 [0.900, 1.072] & 0.938 [0.927, 0.948] \\
\addlinespace
C: non-additive & Full pFedHGP & \textbf{0.229 [0.216, 0.241]} & \textbf{-1.923 [-2.117, -1.729]} & \textbf{0.106 [0.101, 0.111]} & \textbf{0.099 [0.088, 0.110]} & \textbf{0.789 [0.713, 0.866]} & 0.941 [0.932, 0.950] \\
 & No-deviation & 0.242 [0.228, 0.257] & -1.531 [-1.750, -1.311] & 0.112 [0.106, 0.118] & 0.112 [0.101, 0.123] & 0.794 [0.722, 0.867] & 0.938 [0.929, 0.947] \\
 & No-local & 0.375 [0.349, 0.401] & 0.038 [-0.138, 0.214] & 0.176 [0.167, 0.185] & 0.252 [0.212, 0.292] & 1.331 [1.205, 1.457] & 0.945 [0.937, 0.954] \\
 & Global-only & 0.447 [0.426, 0.467] & 3.048 [2.901, 3.195] & 0.305 [0.295, 0.315] & 0.987 [0.846, 1.127] & 1.910 [1.793, 2.027] & 0.943 [0.938, 0.949] \\
 & Local-only & 0.356 [0.339, 0.374] & 1.008 [0.700, 1.316] & 0.167 [0.158, 0.175] & 0.170 [0.157, 0.183] & 1.339 [1.208, 1.469] & 0.940 [0.931, 0.949] \\
 & pFedGP & 0.301 [0.282, 0.320] & -0.630 [-0.921, -0.340] & 0.133 [0.126, 0.140] & 0.159 [0.141, 0.177] & 0.921 [0.838, 1.005] & 0.938 [0.932, 0.945] \\
\addlinespace
D: correlated & Full pFedHGP & \textbf{0.200 [0.193, 0.206]} & \textbf{-2.589 [-2.717, -2.461]} & \textbf{0.094 [0.091, 0.097]} & \textbf{0.088 [0.081, 0.096]} & \textbf{0.677 [0.622, 0.732]} & 0.938 [0.929, 0.947] \\
 & No-deviation & 0.212 [0.205, 0.219] & -2.169 [-2.292, -2.047] & 0.098 [0.096, 0.101] & 0.100 [0.091, 0.109] & 0.714 [0.662, 0.766] & 0.941 [0.932, 0.950] \\
 & No-local & 0.335 [0.308, 0.362] & -0.749 [-0.894, -0.604] & 0.152 [0.144, 0.161] & 0.250 [0.198, 0.303] & 1.115 [1.012, 1.218] & 0.944 [0.936, 0.953] \\
 & Global-only & 0.380 [0.368, 0.392] & 2.546 [2.388, 2.704] & 0.274 [0.264, 0.285] & 1.032 [0.865, 1.199] & 1.545 [1.422, 1.667] & 0.942 [0.936, 0.949] \\
 & Local-only & 0.324 [0.309, 0.339] & 0.303 [0.065, 0.541] & 0.150 [0.143, 0.157] & 0.169 [0.151, 0.187] & 1.172 [1.056, 1.288] & 0.941 [0.932, 0.949] \\
 & pFedGP & 0.266 [0.254, 0.278] & -1.131 [-1.398, -0.863] & 0.117 [0.113, 0.122] & 0.142 [0.129, 0.155] & 0.835 [0.767, 0.904] & 0.940 [0.931, 0.949] \\
\bottomrule
\end{tabular}}
\end{table}

The first pattern is that the full pFedHGP is consistently the strongest specification. In the matched scenario, Full pFedHGP gives the best overall combination of point prediction, probabilistic fit, interval sharpness, and covariance recovery. The ablations show why this improvement is not attributable to a single pooled multi-output GP. Removing the deviation component produces a smaller but systematic degradation, indicating that \(\mathbf{f}_{\delta,i}\) captures smooth client-specific calibration beyond a global-plus-local model. Removing the local component produces a much larger loss in point prediction and probabilistic sharpness, indicating that \(\mathbf{f}_i\) is needed to represent short-scale variation. Global-only underperforms because it cannot personalize to heterogeneous clients, while Local-only and pFedGP improve on Global-only but do not recover the same balance between shared structure, client drift, local variation behavior, and output covariance.

\begin{figure}[ht]
\centering
\includegraphics[width=1\textwidth]{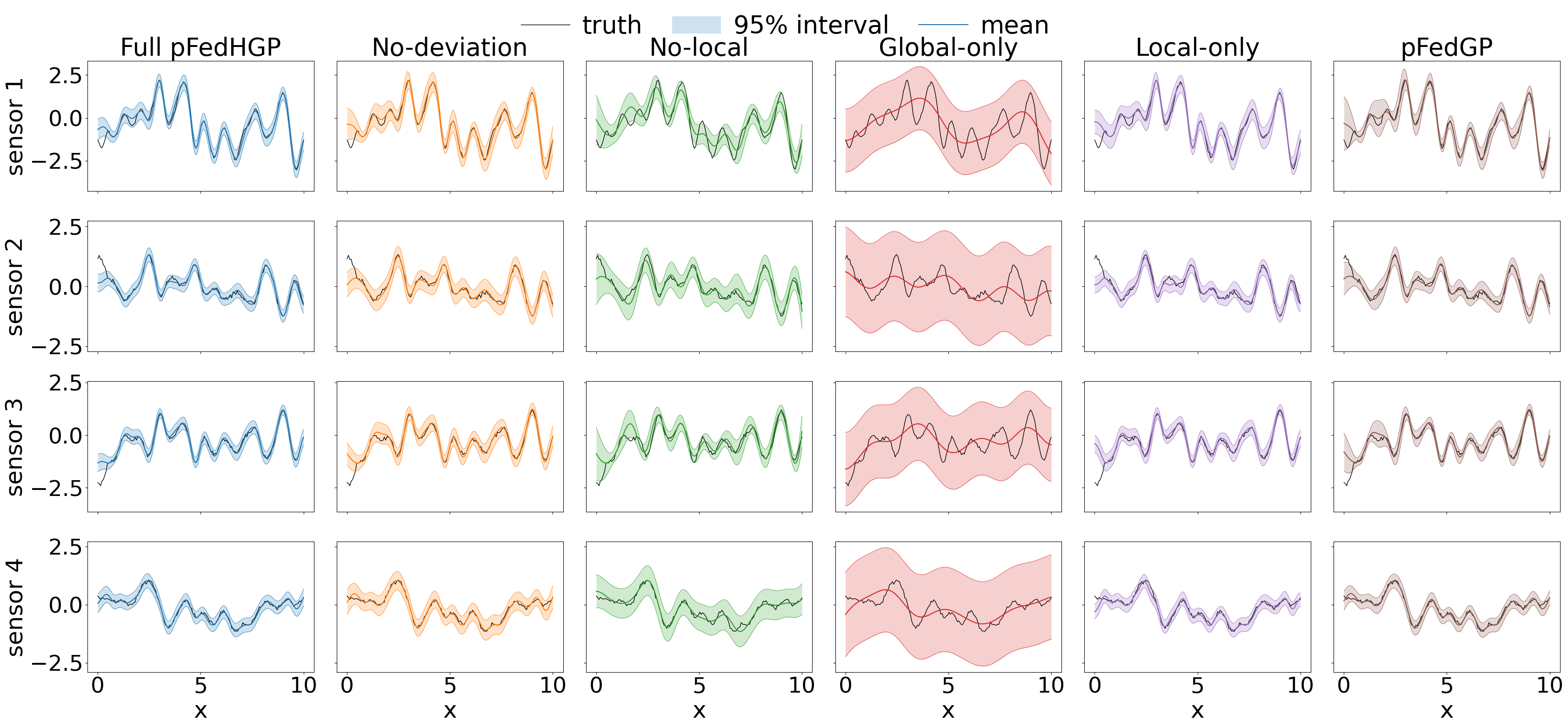}
\caption{Representative four-channel prediction for one client in the matched benchmark. The model jointly predicts correlated outputs and reports 95\% predictive intervals.}
\label{fig:multioutput_prediction}
\end{figure}

Figure~\ref{fig:multioutput_prediction} provides a representative client visualization for the matched multi-output setting. The four channels exhibit different amplitudes and local trajectory patterns, reflecting the cross-output mixing used in the data-generating process. This figure is therefore used as a visual check of the multi-output regression behavior; the recovery of cross-output covariance is evaluated quantitatively by \(\operatorname{CovErr}\) in Table~\ref{tab:robustness}, using the ground-truth and model-implied covariances in~\eqref{eq:true_output_covariance}--\eqref{eq:cov_error} and the fitted ILMM--pFedHGP protocol in Section~\ref{sec:exp_implementation}.

The second pattern is robustness of prediction under misspecification, which we interpret as a representation-efficiency issue rather than a failure of the function class. The fitted layers use squared-exponential kernels, which are universal on compact input domains: continuous targets on the experimental interval can be approximated by an RBF GP even when the true covariance is periodic, the true map is multiplicative, or the latent layers are correlated. Universality means the misspecified targets remain approximable. It does not make those representations equally cheap, and it is not a claim that the true kernel, the multiplicative mechanism, or the independent-layer split is recovered. The hierarchy is an efficient parameterization when its assumptions hold. When they fail, the same universal class can still fit the observable mean and covariance, but it uses extra residual variance, shorter effective local lengthscales, and a finite inducing budget less economically. Point prediction can still be approximated, so RMSE need not collapse; leftover structure is pushed into residual variance, which is why NLL worsens and intervals widen.

This view matches Table~\ref{tab:robustness}. Full pFedHGP remains the best or near-best method in Scenarios~B--D, yet the mismatch is not free. In Scenario~B, RMSE rises from \(0.223\) to \(0.256\), NLL from \(-2.275\) to \(-0.996\), and the average 95\% interval width from \(0.765\) to \(0.919\); coverage stays near nominal because the intervals widen. The periodic deviation is approximable by RBF layers, but a scaled copy of \(k_g\) is an inefficient code for periodicity, so the local residual must absorb the leftover oscillation. Reduced models have fewer universal components and therefore pay a larger efficiency cost: No-local and Global-only fold the unmatched structure into a single kernel. In Scenario~C the product \(\mathbf{f}_g\odot\mathbf{f}_{\delta,i}\) is a continuous function of \(x\) and can be approximated by the remaining RBF layers, especially \(f_i\); the additive split is simply a less efficient code for a multiplicative interaction, which shows up mainly as a worse NLL (\(-2.275\) to \(-1.923\)) rather than a large RMSE gap. In Scenario~D, \(\mathbf{f}_i=\rho\mathbf{f}_{\delta,i}+\sqrt{1-\rho^2}\widetilde{\mathbf{f}}_i\) is already a linear reparameterization of two GPs, so predictive scores need not degrade; the efficiency cost is replaced by an interpretability cost, because the fitted independent-layer split is not the true correlated attribution.

Thus, what remains robust is the ability of a richer universal-kernel hierarchy to approximate the observable predictive distribution. The price of mismatch is representation efficiency: weaker NLL and wider intervals when the deviation kernel is wrong, and a weaker literal reading of \(f_g\), \(f_{\delta,i}\), and \(f_i\) when additivity or independence fails.

\begin{figure}[ht]
\centering
\includegraphics[width=1\textwidth]{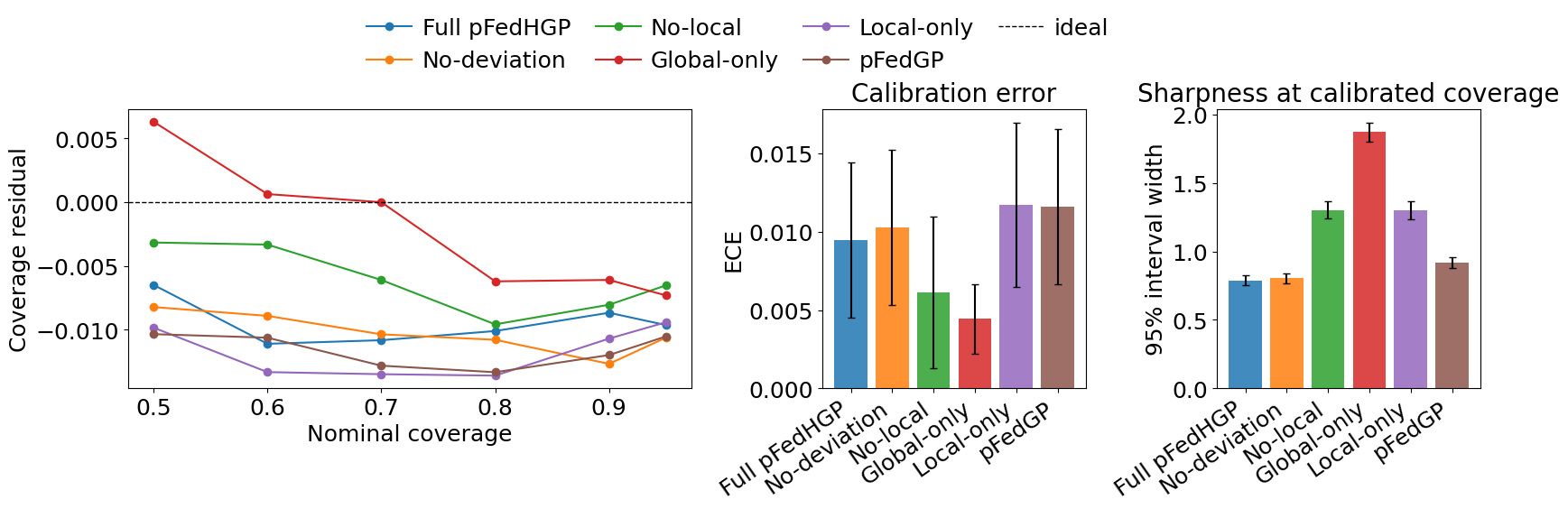}
\caption{Uncertainty calibration and sharpness diagnostics for the multi-output synthetic benchmark. The left panel plots empirical coverage minus nominal coverage, with the dashed horizontal line denoting ideal calibration. The middle panel reports expected calibration error (ECE), and the right panel reports the average 95\% interval width.}
\label{fig:reliability_diagram}
\end{figure}

The third pattern concerns uncertainty calibration and sharpness. Across scenarios, the empirical 95\% coverages are close to the nominal level for all methods. Full pFedHGP is not the closest method to nominal coverage, nor does it have the smallest calibration error. We do not interpret coverage alone as a ranking criterion. Instead, the relevant comparison is the coverage--sharpness tradeoff together with proper probabilistic scores. Figure~\ref{fig:reliability_diagram} makes this tradeoff explicit: the residual calibration curves show that all methods have small deviations from nominal coverage, the ECE panel summarizes the remaining calibration error across nominal levels, and the interval-width panel compares sharpness at comparable calibration. Reduced models such as No-local and Global-only achieve smaller calibration error by producing wider, more conservative intervals, reflecting their need to absorb unmodeled client heterogeneity into predictive variance. In contrast, the full pFedHGP explicitly represents shared global structure,  client-specific deviation, and local site-specific variation, leading to sharper intervals while remaining close to nominal coverage. Together with the RMSE, NLL, CRPS, CovErr, and width columns in Table~\ref{tab:robustness}, these diagnostics show that pFedHGP provides a more favorable probabilistic prediction tradeoff.

\section{Application Case Studies}
\label{sec:case study}

\subsection{Manufacturing Monitoring: Press Tonnage and Fault Diagnosis}
\label{sec:tonnage}

\begin{figure}[ht]
\centering
\includegraphics[scale=0.6]{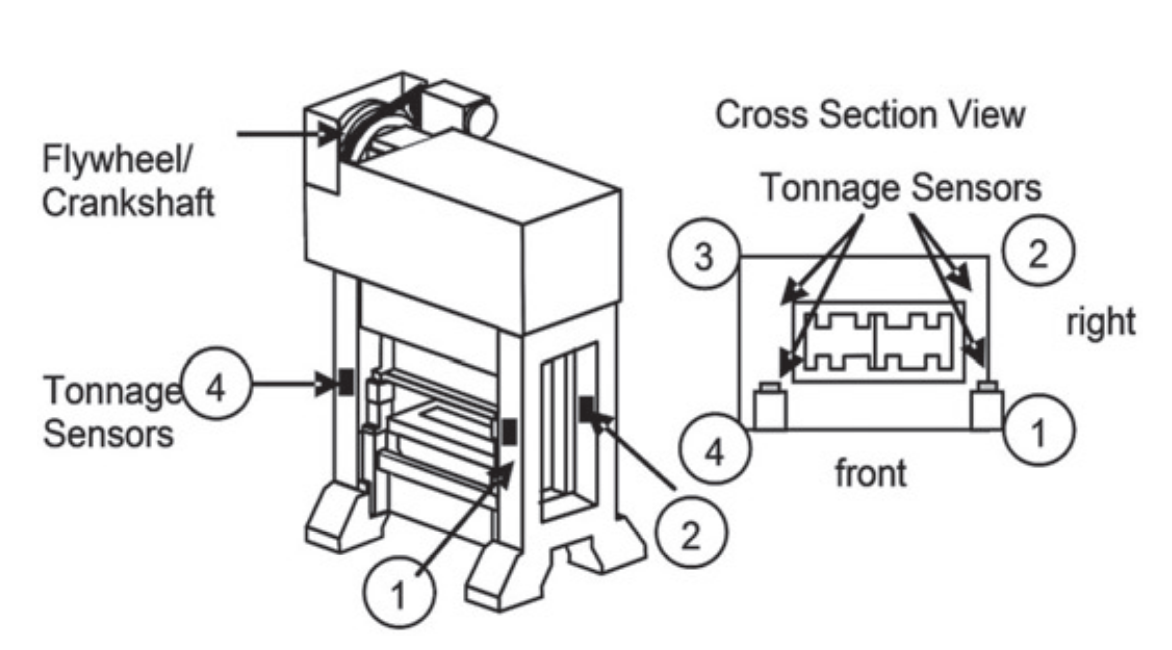}
\caption{Schematic of tonnage monitoring system with four strain-gauge sensors on the press uprights.}
\label{fig:tonnage_create}
\end{figure}

This case study evaluates pFedHGP as a federated personalized Gaussian process model for real-time monitoring and fault diagnosis of a multi-station metal stamping press. The physical system is a multi-operation forming line where billets pass through sequential stamping stations. Four strain-gauge tonnage sensors are mounted on the press uprights, as illustrated in Figure~\ref{fig:tonnage_create}. Each production cycle generates a four-channel force--time waveform that serves as the observed monitoring trajectory \citep{lei2010automatic}. The prediction task is to distinguish normal operation from four distinct fault modes, corresponding to missing parts at different stations, while maintaining calibration to the shared physical press dynamics. The manufacturing data provide a high-dimensional waveform benchmark with shared process structure and condition-specific variation. Since all cycles are generated by the same press process, the waveforms contain common mechanical patterns induced by the material, tooling, and forming sequence. Missing-part conditions alter these patterns in different local regions of the waveform, making the task suitable for evaluating whether pFedHGP can learn shared structure while adapting to condition-specific behavior.

\begin{figure}[ht]
\centering
\includegraphics[scale=0.5]{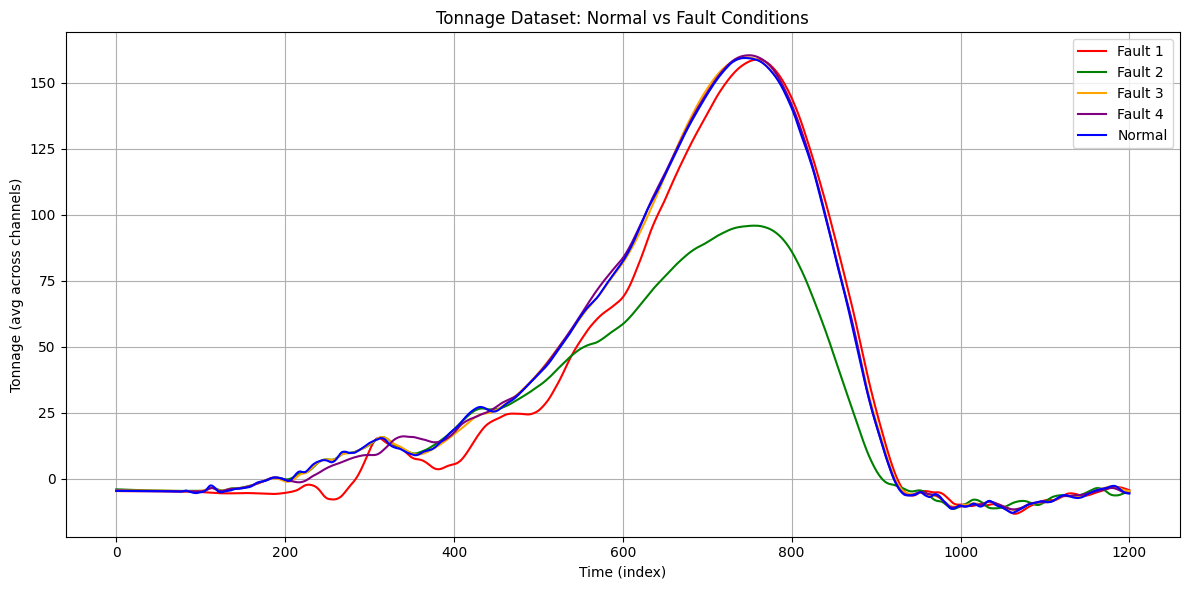}
\caption{Average tonnage waveforms under normal operation and four fault conditions, with shaded bands denoting standard deviation across samples.}
\label{fig:tonnage_curve}
\end{figure}

The dataset comprises $305$ normal cycles and $69$ cycles for each of four fault classes, with $1201$ time points per channel. Thus, the normal class has shape $4\times 1201\times 305$, and each fault class has shape $4\times 1201\times 69$. Representative waveforms for the normal and four fault conditions are shown in Figure~\ref{fig:tonnage_curve}. The five clients in this federated benchmark are therefore the normal class and the four fault classes. These clients correspond to operating conditions in the available single-press dataset, rather than physically isolated factories or production lines.

\paragraph{Why federation is relevant here.}
Although the available tonnage dataset is collected from a single press system, it provides a prototype for distributed manufacturing monitoring settings in which waveform repositories are generated and owned locally by different machines, lines, plants, suppliers, or production units. Raw tonnage trajectories may reveal machine condition, tooling configuration, process settings, and fault signatures, and therefore may be difficult to pool centrally. The high dimensionality of each cycle, with four channels and $1201$ time points, also creates communication and storage costs if raw trajectories are continuously transmitted to a central server. pFedHGP addresses this deployment setting by performing posterior updates locally and communicating model-level updates tied to the shared representation, while retaining client-specific deviation and local residual structure on private data.

For data preparation, each cycle is represented directly by its raw four-channel time indices without filtering or resampling. Channels are concatenated at the cycle level to form the input representation. The training set includes the first $40$ normal samples and $10$ samples per fault class, while the test set includes the remaining $265$ normal and $59$ fault samples per class. This corresponds to using only $13.77\%$ of the available data for training.

The pFedHGP model is configured with a shared inducing grid \(\mathbf{Z}_g\) of $M=81$ points, coupled with a squared-exponential kernel parameterized by learnable lengthscale and variance. Deviations \(\delta_i\) for each client are defined on the same global basis, with scaling parameter \(\phi\) modulating their amplitude relative to the global latent vector \(\mathbf{u}_g\). Each client also maintains a local inducing grid \(\mathbf{Z}_i\) of $M_i=41$ points, with its own squared-exponential kernel to capture higher-frequency, class-specific variations. Initial hyperparameters reflect the observed temporal scales: global lengthscale $50$, local lengthscale $10$, and output variances $1.0$ and $0.5$, respectively. Training proceeds for 10 federated rounds. In each round, clients perform 100 local ELBO steps using Adam with learning rate $0.1$, after which the server aggregates client-wise global gradient contributions and broadcasts the updated global block. Inducing locations are learnable and updated jointly with variational parameters.

At inference, given a test cycle \((\mathbf{X}_\star,\mathbf{y}_\star)\), client \(i\) produces the Gaussian predictive law in Section~\ref{sec:classification}, with mean \(\boldsymbol{\mu}_{\mathbf{f}_i}(\mathbf{X}_\star)\) and full covariance \(\boldsymbol{\Sigma}_i(\mathbf{X}_\star)\) given by~\eqref{eq:pred_cov_classification}. Classification uses the predictive log-likelihood~\eqref{eq:cls-ll} and the maximum-evidence rule~\eqref{eq:cls-rule} over client models \(\{\text{normal},\text{fault 1},\dots,\text{fault 4}\}\).

We compare pFedHGP with established manufacturing monitoring approaches, including personalized Tucker decomposition \citep{hu2025personalized}, hierarchical PCA pipelines \citep{lei2010automatic}, tensor discriminant learning \citep{ye2019monitoring}, CNNs with multilinear PCA features \citep{guo2022multi}, and recurrence-plot imaging \citep{zhou2015automatic}. For pFedHGP, we repeat the experiment five times with different random seeds, and report the resulting mean accuracy and standard deviation. Because the competing methods were reported in separate studies and their full implementations are not available, we treat them as published-reference baselines. Accordingly, Table~\ref{tab:results} reports the training percentages and classification accuracies from the corresponding papers. Standard deviations are included if they are reported in original paper; otherwise, they are omitted because the original papers did not report the necessary variability information.

Table~\ref{tab:results} demonstrates that the proposed federated personalized GP model achieves perfect classification accuracy with only $13.77\%$ of available data, outperforming even perTucker ($15.49\%$ data requirement) and substantially exceeding traditional centralized methods that require $75$--$90\%$ of data. This data efficiency is critical for distributed manufacturing monitoring environments where: (1) fault conditions are rare and labeled training data is expensive, (2) privacy or data-governance constraints may prevent sharing production data across facilities, and (3) edge devices have limited storage and computational resources. The hierarchical decomposition enables the model to rapidly adapt to new fault signatures through the deviation component while maintaining the shared physical press model, supporting proactive maintenance planning and real-time decision support. The probabilistic predictive scores produced by pFedHGP can support risk-aware intervention strategies, although a dedicated calibration analysis for the tonnage case is left for future work.

\begin{table}[ht]
\centering
\caption{Classification accuracy on the tonnage dataset. pFedHGP achieves perfect accuracy with the smallest training fraction. Baseline results are published-reference results reported in the corresponding papers. Standard deviations are included when they can be obtained from the reported accuracy variance.}
\label{tab:results}
\begin{tabular}{|l|c|c|}
\hline
Method & Accuracy & Training percentage \\ \hline
\textbf{pFedHGP} & \(100.00\% \pm 0.00\%\) & 13.77\% \\ \hline
perTucker & 100\% & 15.49\% \\ \hline
PCA+Classifier & \(99.88\% \pm 0.27\%\) & 75\% \\ \hline
UMDA & 99.81\% & 90\% \\ \hline
CNN & $99.85\% \pm 0.49\%$ & 90\% \\ \hline
Recurrent plot & 97.33\% & 80\% \\ \hline
Multi-linear PCA & 96.72\% & 20\% \\ \hline
\end{tabular}
\end{table}


\subsection{Urban Air Quality: Federated Monitoring and Zone Discovery}
\label{sec:climate}

We next evaluate pFedHGP as a federated personalized Gaussian process model for distributed environmental monitoring and urban zone discovery. The study comprises twelve air-quality monitoring stations across Beijing, spanning urban core sites and suburban periphery zones. Each station collects hourly meteorology and pollutant concentrations and is treated as one client maintaining its own local sensor records. The prediction and discovery task is to learn shared regional air-quality structure, capture station-specific variation, identify distinct pollution behaviors across urban zones, and avoid centralized aggregation of station-level time series. To construct a temporal series that captures both diurnal cycles and seasonal trends while maintaining computational efficiency, we subsample measurements at 02:00 and 14:00 daily, yielding approximately 730 observations per year per station. The observed PM2.5 concentration time series is modeled using pFedHGP following the federated training procedure in Section~\ref{sec:methodology}.

\paragraph{Why federation is essential here.}
This air-quality monitoring study directly reflects a distributed sensing setting. Monitoring stations are geographically separated sites, and station-level meteorological and pollutant histories may be owned, managed, or governed by different agencies, municipal units, or local operators. Centralizing all raw station records would require continuously transmitting and storing large volumes of temporal sensor data in a single repository, which creates substantial communication and storage costs.
 pFedHGP addresses this setting by keeping observations at each station, performing local posterior updates on private data, and communicating model-level updates needed to learn the shared regional representation.

\vspace{0.5em}\noindent\textbf{Federated model configuration.}  
The shared component uses $M=30$ inducing points placed uniformly along the annual temporal grid to capture Beijing's regional seasonal pollution dynamics driven by meteorology and emission patterns. Each station maintains a local inducing grid of size $M_i=12$ to model site-specific transient variations from local traffic, industrial sources, and microclimate effects. During federated model synchronization, each station updates its local variational parameters using private sensor streams, while the central server aggregates global parameters representing the shared regional air-quality model. This hierarchical decomposition separates city-wide pollution trends from station- and zone-specific behavior, supporting interpretable urban environmental modeling under data locality.

\vspace{0.5em}\noindent\textbf{Urban zone discovery.}  
After training, each station is characterized by its learned projection operator combining deviation and local components (Section~\ref{sec:usage}), which encodes the station's pollution behavior relative to the shared regional baseline. These operators enable unsupervised discovery of urban zones with similar environmental characteristics. Pairwise dissimilarities computed via Frobenius distance produce a station similarity matrix, which is input to spectral clustering with three groups matching the expected geographical structure: urban core, western suburbs, and eastern suburbs.

\begin{figure}[ht]
\centering
\includegraphics[width=0.7\textwidth]{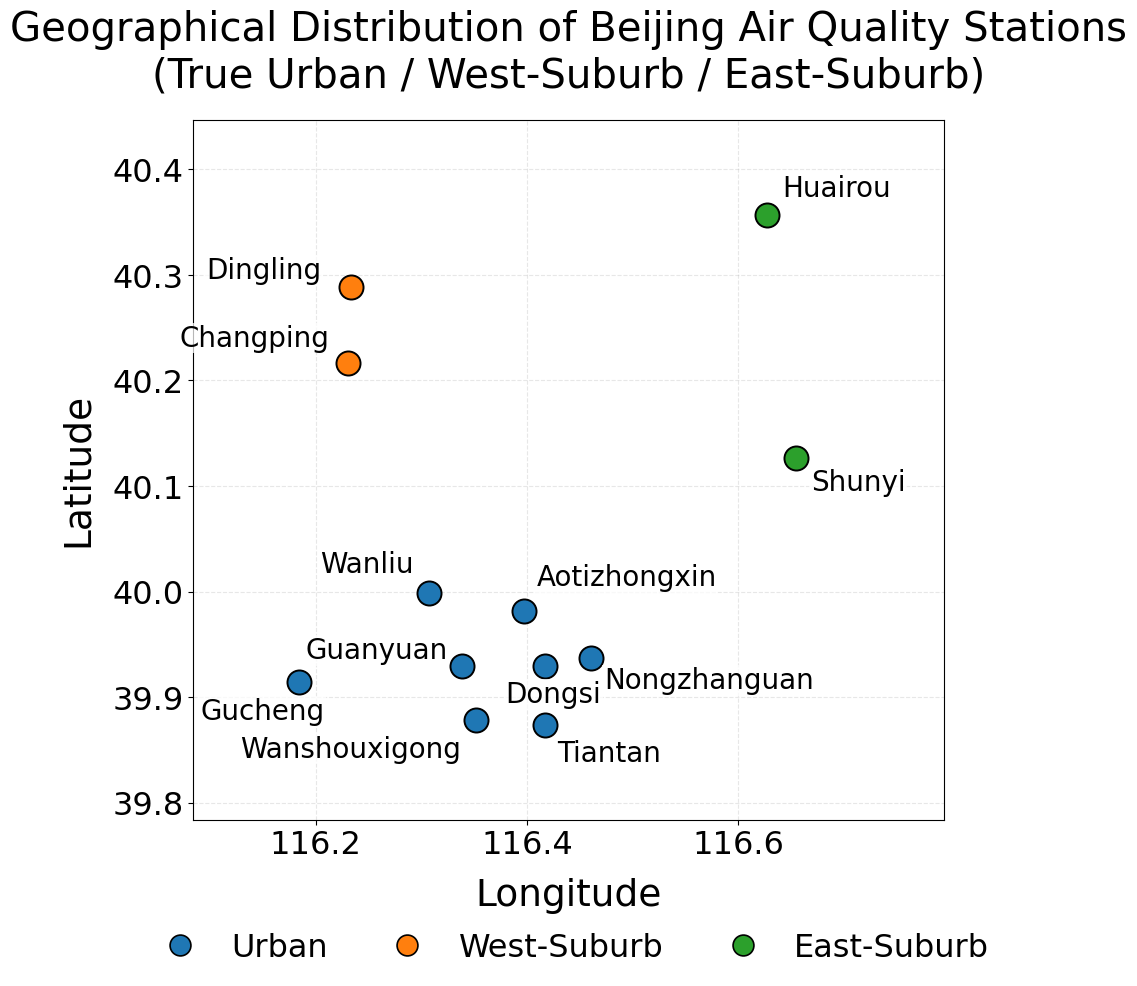}
\caption{True geographical distribution of Beijing monitoring stations. Colors indicate ground-truth categories: Urban (blue), West Suburb (orange), East Suburb (green).}
\label{fig:bj_map}
\end{figure}

\begin{figure}[ht]
\centering
\includegraphics[width=1\textwidth]{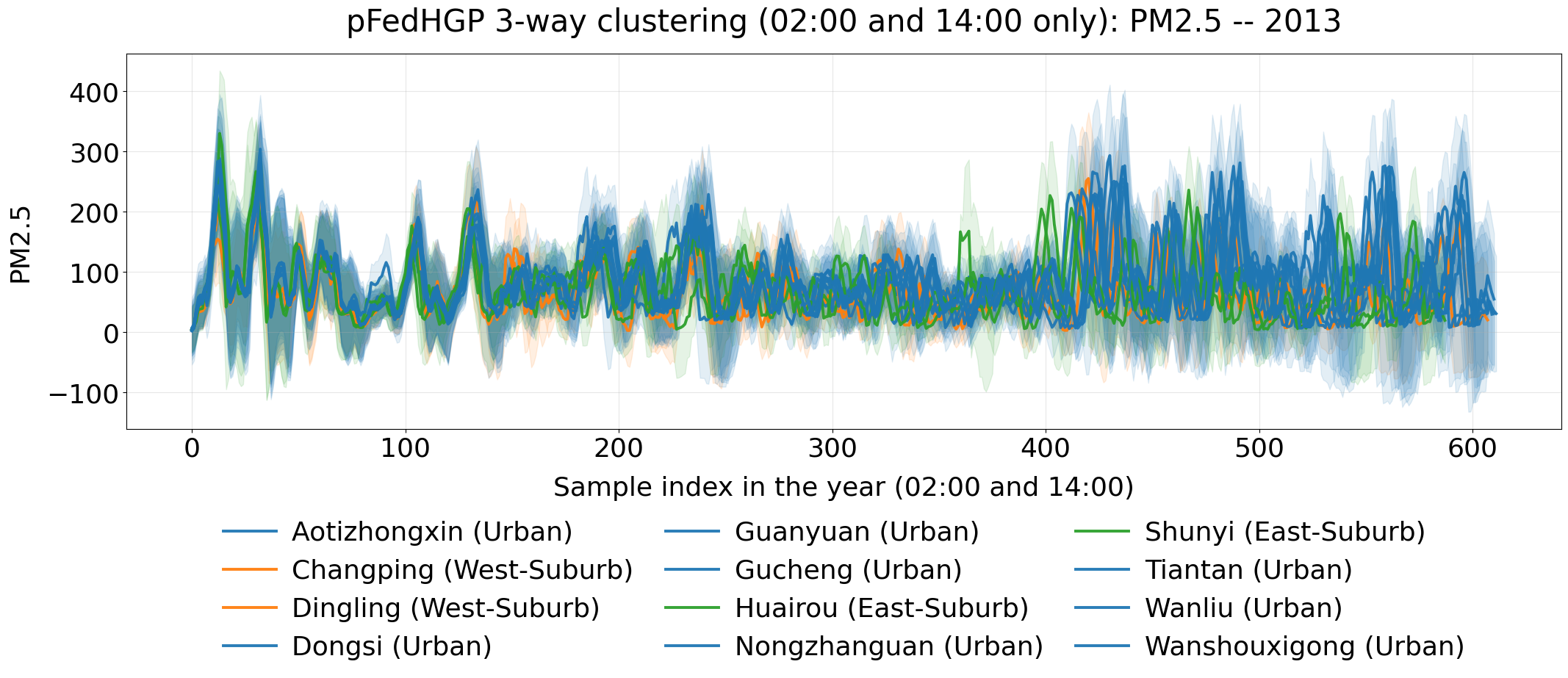}
\caption{pFedHGP clustering result on PM2.5 measurements (2013, 02:00 \& 14:00 only). Colors denote spectral clustering output: Urban (blue), West Suburb (orange), East Suburb (green). The grouping matches the true geographical distribution.}
\label{fig:climate_curve}
\end{figure}

\vspace{0.5em}\noindent\textbf{Results.}  
Figure~\ref{fig:bj_map} shows the ground-truth geographical distribution: urban core (blue), western suburbs including Changping and Dingling (orange), and eastern suburbs including Huairou and Shunyi (green). Figure~\ref{fig:climate_curve} shows the pFedHGP clustering result using 2013 PM2.5 measurements. pFedHGP recovers the three geographical zones: urban stations cluster together despite their physical separation, while western and eastern suburban stations form distinct groups. This result shows that the learned hierarchical representation captures shared regional variation and station-specific pollution behavior without using explicit geographical coordinates or centralizing raw station histories. Such automatic zone discovery supports distributed environmental monitoring tasks including targeted pollution analysis, sensor-network assessment, and air-quality modeling that accounts for both regional patterns and local urban effects. The federated architecture keeps station-level measurements local while still enabling city-wide coordination through shared model updates.

\section{Limitations and Future Work}
\label{sec:limitations}

Despite its advantages, pFedHGP has several limitations. First, the current federated protocol keeps raw observations local but does not by itself provide formal differential privacy or secure aggregation guarantees, so model-update leakage remains a possible risk. Second, robustness to malicious or corrupted clients is not explicitly handled. Third, because $k_\delta=\phi k_g$, the split between $f_g$ and $f_{\delta,i}$ is not uniquely identifiable from a single client's marginal process, and the hierarchical attribution can become less stable when the number of clients is small, when clients have highly unbalanced sample sizes, or when cross-client heterogeneity is weak. Fourth, performance depends on kernel and inducing-budget choices; overly restrictive kernels may underfit systematic deviations, whereas overly flexible local kernels may absorb shared structure. Fifth, communication cost grows with the number of inducing variables and output factors, so adaptive inducing-budget selection is an important direction for larger deployments. Future work will study the conditions under which this hierarchical attribution is stable, including the number of clients, sample-size balance, and degree of cross-client heterogeneity, as well as differentially private updates, online updating, robust aggregation under adversarial participation, and scalable inducing-budget selection in multi-output federated settings.

\section{Conclusion}
\label{sec:conclusion}

We introduced pFedHGP, a federated personalized Gaussian-process framework for probabilistic modeling across heterogeneous clients with data that remain locally held. Its hierarchical construction combines a shared global process, a structured client-specific deviation defined on the shared basis, and a flexible local residual process. This organization supplies a regularized representation of common and client-specific variation while sparse variational inference and federated aggregation allow the shared representation to be learned without centralizing raw observations.

The ILMM formulation places pFedHGP within a standard multi-output GP construction and provides a principled representation of cross-sensor dependence. Our multi-output numerical study evaluates the resulting predictive distributions through point-prediction, proper probabilistic, calibration, sharpness, and output-covariance metrics. Across the matched setting, reduced hierarchical variants, and the specified departures from the fitted assumptions, the full hierarchy provides the most favorable overall predictive trade-off. These results concern recovery of the observable predictive mean and covariance, rather than exact identification of every latent component.

The application studies illustrate complementary distributed prediction tasks. In the manufacturing study, pFedHGP supports fault diagnosis from high-dimensional multi-sensor tonnage waveforms using a limited labeled training set. In the air-quality study, it learns a shared regional representation while retaining station-level records locally and supports the discovery of geographically meaningful station groups without explicit coordinates. Together, these studies demonstrate how federated hierarchical GP modeling can combine information across heterogeneous clients while preserving local adaptation and uncertainty-aware prediction.

\appendix
\section{Derivation of pFedHGP Generative Model}
\label{app:model_derivation}

We detail the construction of the interpolation matrices $\mathbf{A}_{g,i}$ and $\mathbf{A}_{i}$ appearing in Proposition~\ref{prop:low_rank}. The global function $f_g(\cdot)$ and client-specific deviation $f_{\delta,i}(\cdot)$ share the same kernel $k_g$ and inducing inputs $\mathbf{Z}_g$. They are approximated via standard kernel interpolation (sparse GP) formulas:
\[
f_g(\mathbf{x}) = \mathbf{K}_g(\mathbf{x}, \mathbf{Z}_g)\,\mathbf{K}_g(\mathbf{Z}_g, \mathbf{Z}_g)^{-1} \mathbf{u}_g, \quad
f_{\delta,i}(\mathbf{x}) = \mathbf{K}_g(\mathbf{x}, \mathbf{Z}_g)\,\mathbf{K}_g(\mathbf{Z}_g, \mathbf{Z}_g)^{-1} \boldsymbol{\delta}_i.
\]
Summing these yields the total global contribution for client $i$: $f_{g,i}(\mathbf{x}) = \mathbf{K}_g(\mathbf{x}, \mathbf{Z}_g)\,\mathbf{K}_{gg}^{-1} (\mathbf{u}_g + \boldsymbol{\delta}_i)$, where $\mathbf{K}_{gg} = \mathbf{K}_g(\mathbf{Z}_g, \mathbf{Z}_g)$.

Similarly, the local residual $f_i(\cdot)$ uses a client-specific kernel $k_i$ and inducing inputs $\mathbf{Z}_i$: $f_i(\mathbf{x}) = \mathbf{K}_i(\mathbf{x}, \mathbf{Z}_i)\,\mathbf{K}_{ii}^{-1} \mathbf{u}_i$, where $\mathbf{K}_{ii} = \mathbf{K}_i(\mathbf{Z}_i, \mathbf{Z}_i)$.

For a dataset $\mathbf{X}_i$, the projection matrices are:
\[
\mathbf{A}_{g,i} = \mathbf{K}_g(\mathbf{X}_i, \mathbf{Z}_g)\,\mathbf{K}_{gg}^{-1}, \quad
\mathbf{A}_{i} = \mathbf{K}_i(\mathbf{X}_i, \mathbf{Z}_i)\,\mathbf{K}_{ii}^{-1}.
\]
Substituting these into the additive model $y = f_g + f_{\delta,i} + f_i + \varepsilon$ yields Eq.~\eqref{eq:obs_decomposition}.

\section{Proof of Variational Free Energy ELBO (Proposition~\ref{prop:vfe_elbo})}
\label{app:vfe_proof}

We derive the Variational Free Energy (VFE) ELBO following \citet{titsias2009variational}. Starting from the standard GP model with observations $\mathbf{y} = \mathbf{f} + \boldsymbol{\varepsilon}$ where $\boldsymbol{\varepsilon} \sim \mathcal{N}(\mathbf{0}, \sigma^2\mathbf{I})$, we introduce inducing variables $\mathbf{u}$ at locations $\mathbf{Z}$.

The variational distribution factorizes as:
\[
q(\mathbf{f}, \mathbf{u}) = p(\mathbf{f} \mid \mathbf{u}) q(\mathbf{u}), \quad q(\mathbf{u}) = \mathcal{N}(\mathbf{m}, \mathbf{S}),
\]
where $p(\mathbf{f} \mid \mathbf{u}) = \mathcal{N}(\mathbf{K}_{nm}\mathbf{K}_{mm}^{-1}\mathbf{u}, \mathbf{K}_{nn} - \mathbf{Q}_{nn})$ is the exact conditional distribution.

The ELBO is defined as:
\[
\mathcal{L}_{\mathrm{VFE}} = \mathbb{E}_{q(\mathbf{f}, \mathbf{u})}[\log p(\mathbf{y} \mid \mathbf{f})] - \mathrm{KL}(q(\mathbf{f}, \mathbf{u}) \Vert p(\mathbf{f}, \mathbf{u})).
\]

Expanding the KL divergence term:
\[
\begin{aligned}
\mathrm{KL}(q(\mathbf{f}, \mathbf{u}) \Vert p(\mathbf{f}, \mathbf{u}))
&= \mathrm{KL}(p(\mathbf{f} \mid \mathbf{u}) q(\mathbf{u}) \Vert p(\mathbf{f} \mid \mathbf{u}) p(\mathbf{u})) \\
&= \mathbb{E}_{q(\mathbf{u})}[\mathrm{KL}(p(\mathbf{f} \mid \mathbf{u}) \Vert p(\mathbf{f} \mid \mathbf{u}))] + \mathrm{KL}(q(\mathbf{u}) \Vert p(\mathbf{u})) \\
&= \mathrm{KL}(q(\mathbf{u}) \Vert p(\mathbf{u})),
\end{aligned}
\]
where the first term vanishes since $p(\mathbf{f} \mid \mathbf{u})$ appears in both distributions.

For the expected log-likelihood term, we have:
\[
\begin{aligned}
\mathbb{E}_{q(\mathbf{f}, \mathbf{u})}[\log p(\mathbf{y} \mid \mathbf{f})]
&= \mathbb{E}_{q(\mathbf{u})}\left[\mathbb{E}_{p(\mathbf{f} \mid \mathbf{u})}[\log p(\mathbf{y} \mid \mathbf{f})]\right] \\
&= \mathbb{E}_{q(\mathbf{u})}[\log p(\mathbf{y} \mid \mathbf{u})] - \frac{1}{2\sigma^2}\operatorname{Tr}(\mathbf{K}_{nn} - \mathbf{Q}_{nn}),
\end{aligned}
\]
where the trace term arises from the difference between the full covariance $\mathbf{K}_{nn}$ and the low-rank approximation $\mathbf{Q}_{nn} = \mathbf{K}_{nm}\mathbf{K}_{mm}^{-1}\mathbf{K}_{mn}$ in the conditional distribution $p(\mathbf{f} \mid \mathbf{u})$.

Combining these terms yields the VFE ELBO:
\[
\mathcal{L}_{\mathrm{VFE}} = \mathbb{E}_{q(\mathbf{u})}[\log p(\mathbf{y} \mid \mathbf{u})] - \mathrm{KL}(q(\mathbf{u}) \Vert p(\mathbf{u})) - \frac{1}{2\sigma^2}\operatorname{Tr}(\mathbf{K}_{nn} - \mathbf{Q}_{nn}),
\]
which completes the proof. \Halmos

\section{Proof of ELBO Decomposition (Theorem~\ref{prop:elbo_decomp})}
\label{app:elbo_proof}

We provide the derivation of the ELBO decomposition. The variational lower bound on the marginal likelihood is defined as:
\[
\mathcal{L} = \mathbb{E}_{q}[\log p(\{\mathbf{y}_i\}, \mathbf{u}_g, \{\boldsymbol{\delta}_i\}, \{\mathbf{u}_i\})] - \mathbb{E}_{q}[\log q(\mathbf{u}_g, \{\boldsymbol{\delta}_i\}, \{\mathbf{u}_i\})],
\]
where the expectation is taken with respect to the variational posterior $q(\mathbf{u}_g, \{\boldsymbol{\delta}_i\}, \{\mathbf{u}_i\}) = q(\mathbf{u}_g)\prod_{i=1}^T q(\boldsymbol{\delta}_i)q(\mathbf{u}_i)$.

Substituting the factorized joint distribution $p(\{\mathbf{y}_i\}, \mathbf{u}_g, \{\boldsymbol{\delta}_i\}, \{\mathbf{u}_i\}) = p(\mathbf{u}_g) \prod_{i=1}^T p(\mathbf{y}_i \mid \mathbf{u}_g, \boldsymbol{\delta}_i, \mathbf{u}_i) p(\boldsymbol{\delta}_i) p(\mathbf{u}_i)$, we can expand the first term:
\begin{align*}
\mathbb{E}_{q}[\log p] &= \mathbb{E}_{q}\left[ \log p(\mathbf{u}_g) + \sum_{i=1}^T \left( \log p(\mathbf{y}_i \mid \mathbf{u}_g, \boldsymbol{\delta}_i, \mathbf{u}_i) + \log p(\boldsymbol{\delta}_i) + \log p(\mathbf{u}_i) \right) \right] \\
&= \mathbb{E}_{q(\mathbf{u}_g)}[\log p(\mathbf{u}_g)] + \sum_{i=1}^T \left( \mathbb{E}[\log p(\mathbf{y}_i \mid \mathbf{u}_g, \boldsymbol{\delta}_i, \mathbf{u}_i)] + \mathbb{E}[\log p(\boldsymbol{\delta}_i)] + \mathbb{E}[\log p(\mathbf{u}_i)] \right).
\end{align*}

Similarly, the entropy term expands as:
\begin{align*}
\mathbb{E}_{q}[\log q] &= \mathbb{E}_{q}\left[ \log q(\mathbf{u}_g) + \sum_{i=1}^T \left( \log q(\boldsymbol{\delta}_i) + \log q(\mathbf{u}_i) \right) \right] \\
&= \mathbb{E}_{q(\mathbf{u}_g)}[\log q(\mathbf{u}_g)] + \sum_{i=1}^T \left( \mathbb{E}[\log q(\boldsymbol{\delta}_i)] + \mathbb{E}[\log q(\mathbf{u}_i)] \right).
\end{align*}

Combining terms by variable groups:
\begin{itemize}
    \item Global terms: $\mathbb{E}_{q(\mathbf{u}_g)}[\log p(\mathbf{u}_g)] - \mathbb{E}_{q(\mathbf{u}_g)}[\log q(\mathbf{u}_g)] = -\mathrm{KL}(q(\mathbf{u}_g)\|p(\mathbf{u}_g))$.
    \item Local terms for client $i$: 
    \begin{align*}
    &\mathbb{E}[\log p(\mathbf{y}_i \mid \mathbf{u}_g, \boldsymbol{\delta}_i, \mathbf{u}_i)] + (\mathbb{E}[\log p(\boldsymbol{\delta}_i)] - \mathbb{E}[\log q(\boldsymbol{\delta}_i)]) + (\mathbb{E}[\log p(\mathbf{u}_i)] - \mathbb{E}[\log q(\mathbf{u}_i)]) \\
    &= \mathbb{E}[\log p(\mathbf{y}_i \mid \mathbf{u}_g, \boldsymbol{\delta}_i, \mathbf{u}_i)] - \mathrm{KL}(q(\boldsymbol{\delta}_i)\|p(\boldsymbol{\delta}_i)) - \mathrm{KL}(q(\mathbf{u}_i)\|p(\mathbf{u}_i)).
    \end{align*}
\end{itemize}

Summing these components yields the result in Theorem~\ref{prop:elbo_decomp}:
\[
\mathcal{L} = \sum_{i=1}^T \underbrace{\left( \mathbb{E}[\log p(\mathbf{y}_i \mid \mathbf{u}_g, \boldsymbol{\delta}_i, \mathbf{u}_i)] - \mathrm{KL}(q(\boldsymbol{\delta}_i)\|p(\boldsymbol{\delta}_i)) - \mathrm{KL}(q(\mathbf{u}_i)\|p(\mathbf{u}_i)) \right)}_{\mathcal{L}_i} - \mathrm{KL}(q(\mathbf{u}_g)\|p(\mathbf{u}_g)).
\]

\paragraph{Expected log-likelihood with variational covariance.}
For client $i$, the conditional likelihood $p(\mathbf{y}_i \mid \mathbf{u}_g,\boldsymbol{\delta}_i,\mathbf{u}_i)=\mathcal{N}(\mathbf{A}_{g,i}(\mathbf{u}_g+\boldsymbol{\delta}_i)+\mathbf{A}_{i}\mathbf{u}_i,\sigma^2\mathbf{I})$ is linear in the inducing variables. With $q(\mathbf{u}_g)=\mathcal{N}(\mathbf{m}_g,\mathbf{S}_g)$, $q(\boldsymbol{\delta}_i)=\mathcal{N}(\mathbf{m}_{\delta_i},\mathbf{S}_{\delta_i})$, and $q(\mathbf{u}_i)=\mathcal{N}(\mathbf{m}_{u_i},\mathbf{S}_{u_i})$, the predictive mean is $\boldsymbol{\mu}_{\mathbf{f}_i}$ in~\eqref{eq:mean_decomposition} and the extra expected-log-likelihood terms are
\[
-\frac{1}{2\sigma^2}\Bigl(
\operatorname{Tr}(\mathbf{A}_{g,i}\mathbf{S}_g\mathbf{A}_{g,i}^\top)
+\operatorname{Tr}(\mathbf{A}_{\delta,i}\mathbf{S}_{\delta_i}\mathbf{A}_{\delta,i}^\top)
+\operatorname{Tr}(\mathbf{A}_{i}\mathbf{S}_{u_i}\mathbf{A}_{i}^\top)\Bigr),
\]
in addition to the residual trace $\operatorname{Tr}(\mathbf{R}_i)$ from the sparse GP approximation. This yields~\eqref{eq:closed_form_ll} and the explicit ELBO in Theorem~\ref{prop:elbo_decomp}.

\section{Proof of Proposition~\ref{prop:low_rank} (Low-Rank Hierarchical Representation)}
\label{app:proof_prop2}

We prove that the conditional mean of the total latent function admits the low-rank basis expansion stated in Proposition~\ref{prop:low_rank}.

\paragraph{Setup.} 
Under the GP priors, each component process is a Gaussian process:
\[
f_g(\cdot) \sim \mathcal{GP}(0, k_g(\cdot,\cdot)), \quad
f_{\delta,i}(\cdot) \sim \mathcal{GP}(0, \phi k_g(\cdot,\cdot)), \quad
f_i(\cdot) \sim \mathcal{GP}(0, k_i(\cdot,\cdot)).
\]
The inducing variables $\mathbf{u}_g = f_g(\mathbf{Z}_g)$, $\boldsymbol{\delta}_i = f_{\delta,i}(\mathbf{Z}_g)$, and $\mathbf{u}_i = f_i(\mathbf{Z}_i)$ follow the joint Gaussian distributions specified in the proposition.

\paragraph{Conditional Mean Derivation.}
By properties of Gaussian processes, the conditional distribution of $f_g(\mathbf{x})$ given $\mathbf{u}_g$ is Gaussian with mean
\[
\mathbb{E}[f_g(\mathbf{x}) \mid \mathbf{u}_g] = \mathrm{Cov}(f_g(\mathbf{x}), \mathbf{u}_g) \mathrm{Var}(\mathbf{u}_g)^{-1} \mathbf{u}_g = \mathbf{k}_g(\mathbf{x}, \mathbf{Z}_g)\T \mathbf{K}_{gg}^{-1} \mathbf{u}_g,
\]
where $\mathbf{k}_g(\mathbf{x}, \mathbf{Z}_g) = [k_g(\mathbf{x}, \mathbf{z}_{g,1}), \ldots, k_g(\mathbf{x}, \mathbf{z}_{g,M})]^\top$ is the cross-covariance vector.

Similarly, for the client-specific global component: $\mathbb{E}[f_{\delta,i}(\mathbf{x}) \mid \boldsymbol{\delta}_i] = \mathbf{k}_g(\mathbf{x}, \mathbf{Z}_g)\T \mathbf{K}_{gg}^{-1} \boldsymbol{\delta}_i$, noting that $f_{\delta,i}$ uses the same kernel structure as $f_g$.

For the local component: $\mathbb{E}[f_i(\mathbf{x}) \mid \mathbf{u}_i] = \mathbf{k}_i(\mathbf{x}, \mathbf{Z}_i)\T \mathbf{K}_{ii}^{-1} \mathbf{u}_i$.

\paragraph{Total Mean.}
The total latent function is $f_i^{\mathrm{tot}}(\mathbf{x}) = f_g(\mathbf{x}) + f_{\delta,i}(\mathbf{x}) + f_i(\mathbf{x})$. Since the inducing variables from different components are independent, the conditional expectation of the sum equals the sum of conditional expectations:
\[
\begin{aligned}
\mathbb{E}[f_i^{\mathrm{tot}}(\mathbf{x}) \mid \mathbf{u}_g, \boldsymbol{\delta}_i, \mathbf{u}_i] 
&= \mathbb{E}[f_g(\mathbf{x}) \mid \mathbf{u}_g] + \mathbb{E}[f_{\delta,i}(\mathbf{x}) \mid \boldsymbol{\delta}_i] + \mathbb{E}[f_i(\mathbf{x}) \mid \mathbf{u}_i] \\
&= \mathbf{k}_g(\mathbf{x}, \mathbf{Z}_g)\T \mathbf{K}_{gg}^{-1} (\mathbf{u}_g + \boldsymbol{\delta}_i) + \mathbf{k}_i(\mathbf{x}, \mathbf{Z}_i)\T \mathbf{K}_{ii}^{-1} \mathbf{u}_i.
\end{aligned}
\]

For the vectorized form at observations $\mathbf{X}_i$, define $\mathbf{K}_g(\mathbf{X}_i, \mathbf{Z}_g) \in \mathbb{R}^{n_i \times M}$ as the matrix with entries $[\mathbf{K}_g(\mathbf{X}_i, \mathbf{Z}_g)]_{jm} = k_g(\mathbf{x}_{i,j}, \mathbf{z}_{g,m})$. Then:
\[
\begin{aligned}
\mathbb{E}[\mathbf{f}_i^{\mathrm{tot}} \mid \mathbf{u}_g, \boldsymbol{\delta}_i, \mathbf{u}_i] 
&= \mathbf{K}_g(\mathbf{X}_i, \mathbf{Z}_g) \mathbf{K}_{gg}^{-1} (\mathbf{u}_g + \boldsymbol{\delta}_i) \\
&\quad + \mathbf{K}_i(\mathbf{X}_i, \mathbf{Z}_i) \mathbf{K}_{ii}^{-1} \mathbf{u}_i,
\end{aligned}
\]
which is the low-rank representation stated in~\eqref{eq:low_rank_expansion}. \Halmos

\section{Proof of Proposition~\ref{prop:covariance} (Hierarchical Conditional Covariance Decomposition)}
\label{app:proof_prop3}

We prove the additive decomposition of the conditional covariance stated in Proposition~\ref{prop:covariance}.

\paragraph{Conditional Covariance Formula.}
For a Gaussian process $f \sim \mathcal{GP}(0, k)$ with inducing variables $\mathbf{u} = f(\mathbf{Z})$, the conditional covariance at observations $\mathbf{X}$ is given by:
\[
\mathrm{Cov}(f(\mathbf{X}) \mid \mathbf{u}) = k(\mathbf{X}, \mathbf{X}) - k(\mathbf{X}, \mathbf{Z}) k(\mathbf{Z}, \mathbf{Z})^{-1} k(\mathbf{Z}, \mathbf{X}).
\]
This is the residual covariance after conditioning on the inducing points.

\paragraph{Component-wise Decomposition.}
For the fixed global component: $\mathrm{Cov}(f_g(\mathbf{X}_i) \mid \mathbf{u}_g) = \mathbf{K}_{g,ii} - \mathbf{Q}_{g,ii}$, where $\mathbf{K}_{g,ii} = k_g(\mathbf{X}_i, \mathbf{X}_i)$ and $\mathbf{Q}_{g,ii} = \mathbf{K}_g(\mathbf{X}_i, \mathbf{Z}_g) \mathbf{K}_{gg}^{-1} \mathbf{K}_g(\mathbf{Z}_g, \mathbf{X}_i)$.

For the client-specific deviation component with $k_\delta=\phi k_g$: $\mathbf{K}_{\delta,ii}=\phi\mathbf{K}_{g,ii}$ and $\mathbf{Q}_{\delta,ii}=\phi\mathbf{Q}_{g,ii}$, so $\mathrm{Cov}(f_{\delta,i}(\mathbf{X}_i) \mid \boldsymbol{\delta}_i) = \mathbf{K}_{\delta,ii} - \mathbf{Q}_{\delta,ii} = \phi(\mathbf{K}_{g,ii} - \mathbf{Q}_{g,ii})$. The scale $\phi$ therefore enters the residual covariance of the deviation layer even though the conditional mean shares the same interpolation matrix as $f_g$.

For the local component: $\mathrm{Cov}(f_i(\mathbf{X}_i) \mid \mathbf{u}_i) = \mathbf{K}_{i,ii} - \mathbf{Q}_{i,ii}$, where $\mathbf{K}_{i,ii} = k_i(\mathbf{X}_i, \mathbf{X}_i)$ and $\mathbf{Q}_{i,ii} = \mathbf{K}_i(\mathbf{X}_i, \mathbf{Z}_i) \mathbf{K}_{ii}^{-1} \mathbf{K}_i(\mathbf{Z}_i, \mathbf{X}_i)$.

\paragraph{Independence and Additivity.}
Under the mean-field variational approximation, the three components $f_g$, $f_{\delta,i}$, and $f_i$ are conditionally independent given their respective inducing variables. Therefore, for the total latent function $\mathbf{f}_i^{\mathrm{tot}} = f_g(\mathbf{X}_i) + f_{\delta,i}(\mathbf{X}_i) + f_i(\mathbf{X}_i)$:
\[
\begin{aligned}
\mathrm{Cov}(\mathbf{f}_i^{\mathrm{tot}} \mid \mathbf{u}_g, \boldsymbol{\delta}_i, \mathbf{u}_i)
&= \mathrm{Cov}(f_g(\mathbf{X}_i) \mid \mathbf{u}_g) + \mathrm{Cov}(f_{\delta,i}(\mathbf{X}_i) \mid \boldsymbol{\delta}_i) \\
&\quad + \mathrm{Cov}(f_i(\mathbf{X}_i) \mid \mathbf{u}_i) \\
&= (\mathbf{K}_{g,ii} - \mathbf{Q}_{g,ii}) + (\mathbf{K}_{\delta,ii} - \mathbf{Q}_{\delta,ii}) + (\mathbf{K}_{i,ii} - \mathbf{Q}_{i,ii}) \\
&= \mathbf{R}_i.
\end{aligned}
\]

\paragraph{Diagonal Residual Structure via Trace Term.}
The VFE objective function (ELBO) contains the expected log-likelihood term $\mathbb{E}_q[\log p(\mathbf{y}|\mathbf{f})]$, which analytically reduces to a form involving $-\frac{1}{2\sigma^2}\operatorname{Tr}(\mathbf{K}_{nn} - \mathbf{Q}_{nn})$. Since the trace operator sums only the diagonal elements, the optimization objective depends only on the marginal variances of the residual process \citep{titsias2009variational}. This allows us to compute the residual term efficiently using only the diagonal elements of $(\mathbf{K}_{\cdot,ii} - \mathbf{Q}_{\cdot,ii})$, without assuming the off-diagonal elements are zero. This yields the computationally efficient form of $\mathbf{R}_i$ mentioned in the proposition. \Halmos

\section{Proof of Proposition~\ref{prop:global_elbo_block} (Global ELBO given fixed local blocks)}
\label{app:global_elbo_block_proof}

Starting from the explicit ELBO decomposition in Theorem~\ref{prop:elbo_decomp},

\begin{align*}
\mathcal{L}
&= \sum_{i=1}^T \Bigl[-\frac{n_i}{2}\log(2\pi \sigma^2)
    - \frac{1}{2\sigma^2}\|\mathbf{y}_i - \boldsymbol{\mu}_{\mathbf{f}_i}\|^2 \\
&\qquad\qquad
    - \frac{1}{2\sigma^2} \bigl(
    \operatorname{Tr}(\mathbf{A}_{g,i}\mathbf{S}_g\mathbf{A}_{g,i}^\top)
    + \operatorname{Tr}(\mathbf{A}_{\delta,i}\mathbf{S}_{\delta_i}\mathbf{A}_{\delta,i}^\top)
    + \operatorname{Tr}(\mathbf{A}_{i}\mathbf{S}_{u_i}\mathbf{A}_{i}^\top) \\
&\qquad\qquad\quad + \operatorname{Tr}(\mathbf{K}_{g,ii} - \mathbf{Q}_{g,ii})
    + \operatorname{Tr}(\mathbf{K}_{\delta,ii} - \mathbf{Q}_{\delta,ii})
    + \operatorname{Tr}(\mathbf{K}_{i,ii} - \mathbf{Q}_{i,ii})\bigr)\Bigr] \\
&\quad - \mathrm{KL}\bigl(q(\mathbf{u}_g)\|p(\mathbf{u}_g)\bigr)
    - \sum_{i=1}^T \mathrm{KL}\bigl(q(\boldsymbol{\delta}_i)\|p(\boldsymbol{\delta}_i)\bigr)
    - \sum_{i=1}^T \mathrm{KL}\bigl(q(\mathbf{u}_i)\|p(\mathbf{u}_i)\bigr),
\end{align*}%

we treat $\{\Theta_{i,\mathrm{local}}^{(r+1)}\}$ as fixed and collect all terms that do not depend on $\Theta_{\mathrm{global}}$ into a constant. The terms depending on $\Theta_{\mathrm{global}}$ are the least squares term, the variational covariance traces $\operatorname{Tr}(\mathbf{A}_{g,i}\mathbf{S}_g\mathbf{A}_{g,i}^\top)$ and $\operatorname{Tr}(\mathbf{A}_{\delta,i}\mathbf{S}_{\delta_i}\mathbf{A}_{\delta,i}^\top)$ (with $\mathbf{A}_{\delta,i}$ depending on $\mathbf{Z}_g$ and $k_g$), the global and deviation residual traces, the global KL, and the deviation KLs $\sum_i \mathrm{KL}(q(\boldsymbol{\delta}_i)\|p(\boldsymbol{\delta}_i;\phi,\mathbf{K}_{gg}))$ because $p(\boldsymbol{\delta}_i)$ depends on $(\phi,\mathbf{K}_{gg})$. This yields \eqref{eq:elbo_global_block}--\eqref{eq:Li_global} up to an additive constant independent of $\Theta_{\mathrm{global}}$. \Halmos

\providecommand{\BIBand}{and}

\end{document}